\documentclass[journal]{vgtc}                

\usepackage[dvipsnames,table]{xcolor}
\usepackage[export]{adjustbox}
\usepackage{appendix}
\usepackage{array}
\usepackage{enumitem}
\usepackage{graphicx}
\usepackage{tikz}
\usepackage{times}
\usepackage{hyperref}
\usepackage{subfig}
\usepackage{amsmath}
\usepackage{amssymb}
\usepackage{algorithm2e}
\usepackage{multirow}
\usepackage{makecell}
\usepackage{float}
\usepackage{multirow}
\usepackage{diagbox}
\usepackage{array}

\usepackage[none]{hyphenat}
\emergencystretch=\maxdimen
\onlineid{0}

\vgtccategory{Research}

\vgtcinsertpkg

\title{The Big Data Myth: Using Diffusion Models for\\ Dataset Generation to Train Deep Detection Models\\ \normalsize{NHL Stenden Professorship in Computer Vision \& Data Science}} 

\author{Roy Voetman \\ \small{Supervisors: Maya Aghaei, Klaas Dijkstra}}
\authorfooter{
\item
  Roy Voetman is a Computer Vision \& Data Science master student at the NHL Stenden University of Applied Sciences, E-mail: roy.voetman@nhlstenden.com.
\item
  Maya Aghaei Gavari is a researcher at the NHL Stenden Professorship in Computer Vision \& Data Science, E-mail: maya.aghaei.gavari@nhlstenden.com.
\item
  Klaas Dijkstra is a professor at the NHL Stenden Professorship in Computer Vision \& Data Science, E-mail: \\klaas.dijksta@nhlstenden.com.
}

\teaser{
  \centering
  \includegraphics[width=0.9\textwidth]{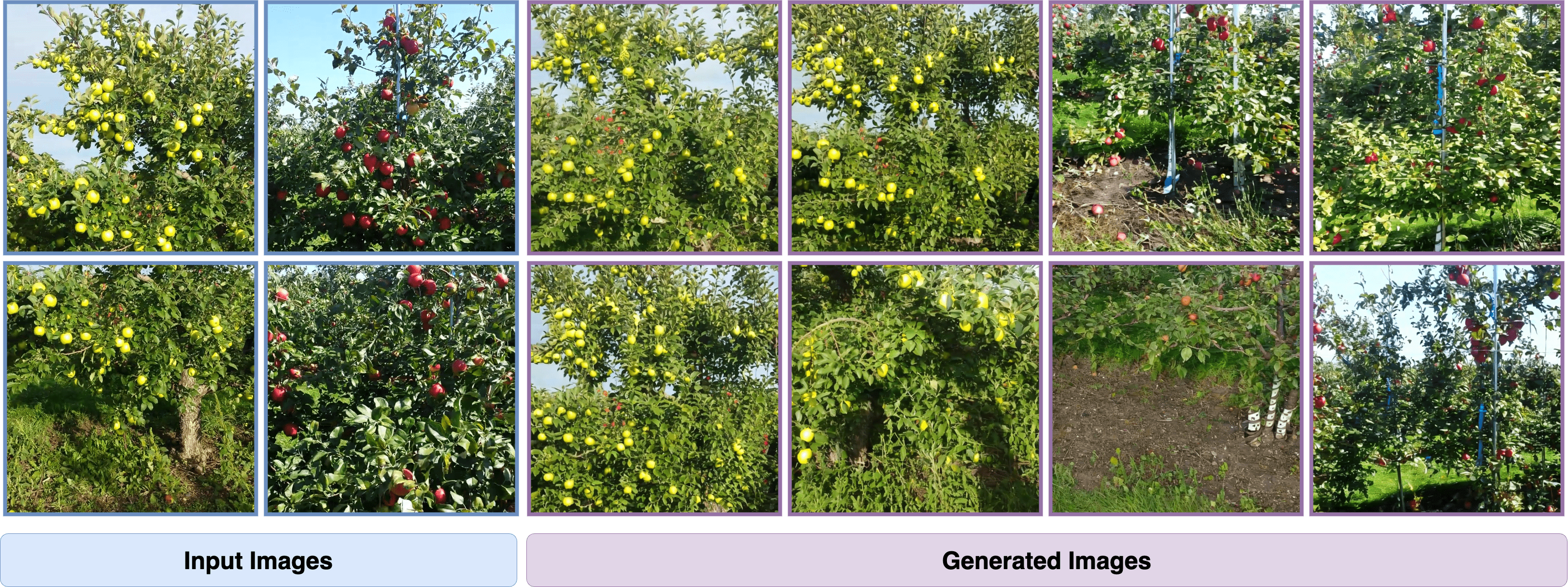}
  \caption{Using 20 real-world images (left) and diffusion models, we generate a dataset (right) to train deep detection models.}
  \label{fig:teaser}
}

\abstract{
Despite the notable accomplishments of deep object detection models, a major challenge that persists is the requirement for extensive amounts of training data. The process of procuring such real-world data is a laborious undertaking, which has prompted researchers to explore new avenues of research, such as synthetic data generation techniques. This study presents a framework for the generation of synthetic datasets by fine-tuning pretrained stable diffusion models. The synthetic datasets are then manually annotated and employed for training various object detection models. These detectors are evaluated on a real-world test set of 331 images and compared against a baseline model that was trained on real-world images. The results of this study reveal that the object detection models trained on synthetic data perform similarly to the baseline model. In the context of apple detection in orchards, the average precision deviation with the baseline ranges from 0.09 to 0.12. This study illustrates the potential of synthetic data generation techniques as a viable alternative to the collection of extensive training data for the training of deep models.} 


\begin{document}

\firstsection{Introduction}

\maketitle

Over the past decade, deep learning has revolutionised the field of computer vision. First, in 2012, the AlexNet \cite{alexnet} architecture sparked a resurgence of interest in Convolutional Neural Networks (CNNs) for image classification. Subsequently, object detection saw advancements through the introduction of Region-based Convolutional Neural Networks (R-CNNs) \cite{rcnn} and You Only Look Once (YOLO) models \cite{yolov1}. Likewise, image segmentation gained popularity with the introduction of SegNet \cite{segnet} and Mask R-CNN \cite{mask-rcnn}. Unfortunately, these models share a common limitation, as their effective training depends heavily on access to a substantial number of images \cite{o2020deep}, often referred to as big data.

Within the computer vision community, considerable efforts are currently being devoted to acquiring representative datasets for training deep models \cite{creating-of-imagenet,sam,bearman2016s,zhang2021cross}. This process involves not only obtaining high-quality images but also the need to manage selection biases, address outliers, and ensure annotation consistency \cite{whang2023data}. Now, suppose it was possible to leverage these deep models effectively without being limited by the availability of extensive datasets. Our paper introduces "Genfusion", a framework that effectively combines recent advances in image synthesis to generate images for training deep models.

Diffusion models \cite{sohl2015deep} have gained recent attention due to their impressive text-to-image synthesis capabilities, as evidenced by notable architectures like DALL-E 2, Stable Diffusion, and Imagen \cite{dall-e-2,stable-diffusion,imagen}. While these models have demonstrated unprecedented results, a significant challenge remains in accurately generating new renditions of identical objects \cite{ruiz2022dreambooth}. However, this capability holds paramount importance for dataset generation where images must closely resemble specific real-world inference scenarios. To address this, we leverage a pretrained diffusion model \cite{stable-diffusion} and employ a fine-tuning technique that operates with a limited set of real-world images \cite{ruiz2022dreambooth}, thereby enabling the generated images to align with a given real-world setting.

To demonstrate the feasibility of this approach, the study takes on the task of apple detection, leveraging a well-established benchmark dataset \cite{hani2020minneapple}. Our results demonstrate the groundbreaking potential of these diffusion models for artificial dataset generation, particularly for challenging and data-scarce real-world applications.

This leads to the following research questions:

\begin{itemize}
  \item How can a pretrained diffusion model be fine-tuned to generate a dataset representative of a specific real-world scenario?
  \item How does the performance of an object detector trained on data generated with \textit{Genfusion} compare to one trained on real data when tested in a real-world scenario?
\end{itemize} 

\subsection{Related work}

Image synthesis is the process of artificially producing images that have a particular desired content. The task is comparable to the inverse of the process of image classification, specifically, the generation of an image with the visual characteristics associated with a particular label. The generation of pictures of non-existing human faces is a well-known example \cite{karras2019style}. The advent of Generative Adversarial Networks (GANs) \cite{goodfellow2020generative} in 2014 marked a significant breakthrough in image synthesis, leading to widespread research in the field \cite{vqgan-clip,tao2022df,zhang2018photographic,hinz2020semantic}. Although recent studies have shown GANs to be successful in generating high-quality images \cite{esser2021taming}, they still face challenges with training stability, convergence, and synthesising diverse datasets with many subject types \cite{de2022fundamentals,diffusion-beats-gan}.

In 2015, diffusion probabilistic models \cite{sohl2015deep} were introduced as another type of generative model. Subsequently, numerous advancements were suggested under the name of denoising diffusion probabilistic models \cite{ho2020denoising, improved-ddpm}, which we will refer to as diffusion models for brevity. Diffusion models have demonstrated high-quality results in image synthesis on datasets with diverse subjects, often outperforming GANs \cite{diffusion-beats-gan}. Conditional image synthesis, a method that generates images based on specific constraints or additional information, is where these models have shown unprecedented performance \cite{dall-e-2, stable-diffusion}. User-provided text prompts are an intuitive way to formulate these constraints, and with the introduction of CLIP \cite{clip}, which encodes natural language in a latent space specifically targeted towards text-to-image correspondence, text-to-image synthesis has become a prominent research direction.

Text-to-image synthesis in general has been researched since 2015, with early proposals utilising Recurrent Neural Networks (RNNs) and Auto Encoders (AEs) \cite{alignDRAW}. In 2021, significant improvements were achieved with VQ-GAN \cite{esser2021taming}, a GAN-based approach, and DALL-E \cite{dall-e}, which applied transformers \cite{transformers} from natural language processing. GLIDE \cite{nichol2021glide} was the first study to integrate text-to-image synthesis with diffusion models, showing even greater diversity in generated images. However, the computational resources required to train the model were extensive, which was a limitation of GLIDE. Shortly after, latent diffusion models \cite{stable-diffusion} were introduced. Inspired by VQ-GAN \cite{esser2021taming} they utilised a Vector Quantized Variational Autoencoder (VQ-VAE) \cite{vq-vae} to perform the diffusion within a compressed latent space, reducing training time. Stable Diffusion is a continuation of the latent diffusion study that incorporates several improvements from related research \cite{github-stable-diffusion}. Stable Diffusion, the only open-sourced pretrained model, will be used for this research, but our approach is not limited to this model.

\subsubsection{Fine-tuning}

There is a relatively small body of literature that is concerned with fine-tuning text-to-image diffusion models for subject-driven generation \cite{gal2022image, gallego2022personalizing, ruiz2022dreambooth}. Subject-driven generation refers to the task of generating new images of a subject in different contexts given only a few images ($\sim$3-5) of that subject. Despite the current outstanding results of state-of-the-art text-to-image models, generating new renditions of the exact same subject, while retaining its critical visual features, is highly challenging \cite{ruiz2022dreambooth}. A recent study proposes a technique to address this challenge via a textual inversion method applied to a pretrained diffusion model \cite{gal2022image}. Textual inversion refers to searching for the optimal embedding to represent visual concepts, such as objects or styles, and connecting this to new pseudo-words (e.g. 'A $S^*$ dog' with $S^*$ denoting the pseudo word). Unfortunately, this approach is limited by the original output domain of the model. Another study proposes a different methodology, namely Aesthetic Gradients, that seeks to personalise the text-to-image model by confining its scope to the domain of aesthetic alterations \cite{gallego2022personalizing}. What sets this approach apart is that the authors retrain the text-encoder exclusively, without necessitating any further fine-tuning of the weights in the diffusion model. However, since this study is limited to aesthetic modifications, it does not offer the generality that is necessary for our research.

A study that stands out from the previously mentioned studies is 'DreamBooth' \cite{ruiz2022dreambooth}. While this study shares similarities with the previously discussed publications in that it employs pseudo-words to describe the subject, the novelty of this method is evident in the fact they \textit{embed} the subject within the output domain of the model via retraining on a small selection of images. This differs from \cite{gal2022image}, which merely \textit{searches} for an embedding in the original output domain that closely resembles the subject. The key challenge with the retraining approach is the issue of overfitting and language drift \cite{lee2019countering,lu2020countering}. This refers to the issue where a language model, which was pretrained on a vast text corpus, experiences a gradual decline in its syntactic and semantic proficiency in the language after fine-tuning for a specific task. This degradation occurs as the model's learning prioritises improvement in the targeted task at the expense of general language knowledge. This is particularly problematic for text-to-image generation, as the model may learn to create images of a specific person, but forget the general concept of a person. The DreamBooth study proposes a \textit{class-specific prior preservation loss} as a regularisation term for the original loss function when fine-tuning for subject-driven generation. This approach incentivises the model, when the pseudo-word is not present, to still produce a variety of distinct images belonging to the same class as our subject.

\subsection{Contribution}

Our work contributes to the field of computer vision by showing that we can combine these techniques into one dataset generation framework. The output of a Stable Diffusion network can be fine-tuned using DreamBooth to replicate images of a specific real-world scenario, which can then be used to train deep learning models. We demonstrate the versatility of this fine-tuning approach by illustrating that it has the capacity to replicate a diverse range of images, rather than solely synthesising images depicting the same object. Additionally, we provide a preliminary analysis on fine-tuning these models for bounding box generation. To demonstrate the feasibility, we will use the YOLOv5 \cite{yolov5} and YOLOv8 \cite{yolov8} object detectors due to their fast inference time as we argue that our use case is for a real-time fruit-picking application. Note that any alternative object detector capable of detecting small objects could be used, but the choice of detector is beyond the scope of this study.

\begin{figure*}[!htb]
\centering
\includegraphics[width=0.8\textwidth]{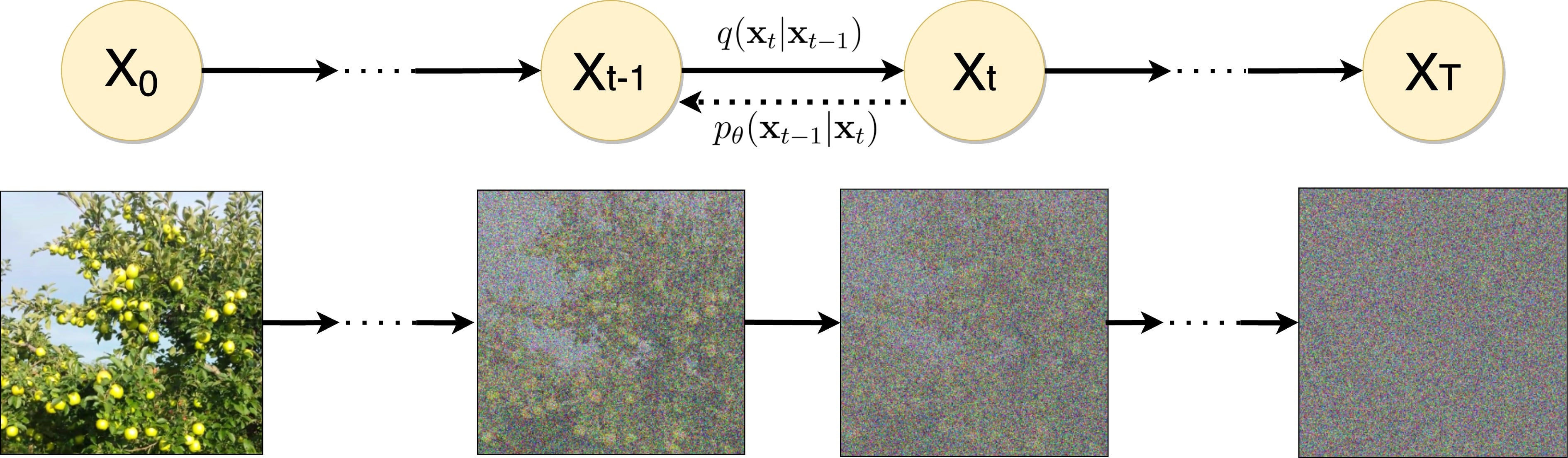}
\caption{The graphical illustration of diffusion models with their fixed forward process $q(\mathbf{x}_t | \mathbf{x}_{t-1})$ and learnt backward process $p_\theta(\mathbf{x}_{t-1} | \mathbf{x}_{t})$. Apple tree illustration adapted from MinneApple \cite{hani2020minneapple}}
\label{fig:forward-process}
\end{figure*}

\section{Methodology} 

To enhance the practicality of training deep object detection models, we observe that the data acquisition process is among the most time-intensive tasks. To mitigate this limitation, we present a multi-stage approach. Initially, we artificially generate training datasets based on a select few real-world images and DreamBooth fine-tuning \cite{ruiz2022dreambooth}. Following this, we manually annotate the generated images and train various YOLOv5 \cite{yolov5} and YOLOv8 \cite{yolov8} object detectors. For evaluation, we test the detectors on the MinneApple apple detection benchmark dataset \cite{hani2020minneapple}. In addition, a preliminary investigation is presented to determine how the diffusion model could possibly be fine-tuned to automatically create annotations.

\subsection{MinneApple Dataset}

The MinneApple dataset \cite{hani2020minneapple} is an apple detection dataset with a collection of 1001 tree images captured in an apple orchard with both green and red apples at varying growth stages, shadings and occlusions. The dataset was published by the University of Minnesota, and data was collected over multiple days to obtain diverse illumination conditions. The image resolution is 1280 × 720 pixels, and the dataset only annotates apples in the foreground while leaving those on the ground and trees in the background unmarked.

To ensure the independence of the test set, the dataset was partitioned into two subsets: a training set consisting of 670 images acquired in 2015, and a testing set consisting of 331 images acquired in 2016. We further divided the training images into a 536-image training set and a 134-image validation set. It is noteworthy that the original training set has an uneven distribution of green and red apples, resulting in our training set having 54 trees with green apples and 482 trees with red apples. Furthermore, in order to conform to the default output size of our Stable Diffusion model, which is 768x768, we have employed a centre cropping technique to create 720x720 cutouts. Subsequently, we have reshaped these cutouts to 768x768 using pixel interpolation.

\subsection{Diffusion Models}

\begin{figure*}[!htb]
    \centering
    \includegraphics[width=1\textwidth]{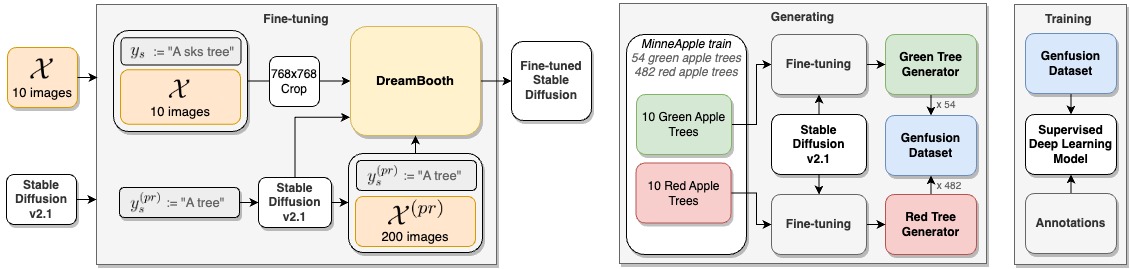}
    \caption{\textbf{The Genfusion Pipeline }\textit{(1) Fine-tuning:} We start with 10 real-world images and a pretrained Stable Diffusion model. These images are subsequently transformed to 768x768 size using centre cropping and pixel interpolation techniques to ensure consistency with the output size of the diffusion model. DreamBooth fine-tuning is then performed on the model using 200 prior preservation images. \textit{(2) Generating:} The fine-tuning is applied twice, once for green apple trees and once for red apple trees, to control the distribution of apple colours in the generated dataset. \textit{ (3) Training:} We use the generated data and annotate it manually with bounding box information to train a deep model.}
    \label{fig:genfusion}
\end{figure*}

Diffusion Models (DMs) \cite{sohl2015deep} are a class of probabilistic models that can be used to model an arbitrary data distribution $q(\mathbf{x})$. The primary observation behind these models is that it is difficult to directly sample new elements from $q(\mathbf{x})$ directly. Instead, diffusion models rely on a so-called diffusion process which gradually transforms $q(\mathbf{x})$ into a noise distribution $\pi(\mathbf{x})$, usually Gaussian $\mathcal{N}(\mathbf{0},\mathbf{I})$, which can be conveniently sampled (see Figure \ref{fig:forward-process}). Subsequently, a function is learnt to invert the noising process to recover the original data distribution $q(\mathbf{x})$. This method is similar to that of autoencoders, with the distinction being the adoption of a fixed encoding process.

More precisely, we define a so-called forward process to transform $q(\mathbf{x})$ into $\pi(\mathbf{x})$ as a Markov Chain of length $T$. The initial state of this chain is our data distribution $q(\mathbf{x})$ denoted as $q(\mathbf{x}_0)$. Each step in the Markov Chain leads to the acquisition of a new distribution $q(\mathbf{x}_t | \mathbf{x}_{t-1})$, which is obtained by adding a small amount of noise to the previous distribution. The final state of the chain, $q(\mathbf{x}_T | \mathbf{x}_{T-1})$, is expected to approximate the noise distribution $\pi(\mathbf{x})$. Equation \ref{eq:forward-step} gives the definition of $q(\mathbf{x}_t | \mathbf{x}_{t-1})$ for Gaussian diffusion, where $\beta_t \in (0, 1)$ is the diffusion rate at time $t$. Intuitively, at $t=0$, the distribution is close to $\mathcal{N}(\mathbf{x}_{t-1}, \mathbf{0})$, indicating that it is based on the prior distribution with very little additional noise. At the other extreme, at time $t=T$, the distribution approaches $\mathcal{N}(\mathbf{0}, \mathbf{I})$, implying that it converges to our unit Gaussian. This convergence is facilitated by the gradual increase in diffusion rate $\beta_t$ over time. The complete forward process that converts the data distribution into the noise distribution is given by Equation \ref{eq:forward-process}.

\begin{equation}
    \label{eq:forward-step}    
    q(\mathbf{x}_{t} | \mathbf{x}_{t-1}) := \mathcal{N}(\mathbf{x}_{t-1}\sqrt{1-\beta_t},\,\mathbf{I}\beta_t)
\end{equation}

\begin{equation}
    \label{eq:forward-process}
    q(\mathbf{x}_{0 \cdots T}) := q(\mathbf{x}_0) \prod_{t=1}^{T} q(\mathbf{x}_{t} | \mathbf{x}_{t-1})
\end{equation}

A reserve process is tasked to learn a function $p_\theta(\mathbf{x}_{t-1} | \mathbf{x}_{t})$ to reverse the noising process. The complete reserve process can be described using Equation \ref{eq:backward-process}. Note that $\theta$ is fixed in time so the weights are time invariant. When training has been completed, the backward process, $p_\theta(\mathbf{x}_{0 \cdots T})$, can produce new samples of $q(\mathbf{x})$ based on sampled Gaussian noise, and the forward process may be disregarded.  

\begin{align}
\label{eq:backward-process}
p(\mathbf{x}_{T}) := \mathcal{N}(\mathbf{0},\,\mathbf{I}), && p_\theta(\mathbf{x}_{0 \cdots T}) := p(\mathbf{x}_T) \prod_{t=1}^{T} p_\theta(\mathbf{x}_{t-1} | \mathbf{x}_{t})
\end{align}

Image synthesis studies commonly implement $p_\theta(\mathbf{x}_{t-1} | \mathbf{x}_{t})$ with a time-conditional UNet \cite{ho2020denoising,improved-ddpm,diffusion-beats-gan,stable-diffusion,unet} (see Appendix \ref{appendix:a}). The model is trained using a loss function that corresponds to a reweighted version of the variation lower bound \cite{ho2020denoising}. In this context, repeatably applying $p_\theta(\mathbf{x}_{t-1} | \mathbf{x}_{t})$ can be interpreted as a sequence of $T$ equally weighted denoising autoencoders with the objective to predict the noise that was added to $\mathbf{x}_0$ given $\mathbf{x}_t$. As the UNet is tasked to predict noise, it is typically denoted as $\epsilon_\theta(\mathbf{x}_t, t)$ which allows us to write the loss function as in Equation \ref{eq:loss-function-dm}. Consequently, the reserve process can be defined as a subtraction of the predicted noise from $\mathbf{x}_{t}$.

\begin{equation}
    \label{eq:loss-function-dm}
    L_{DM} := \mathbb{E}_{\mathbf{x}, \epsilon \sim \mathcal{N}(\mathbf{0}, \mathbf{I}), t} \Bigr[ \lVert \epsilon - \epsilon_\theta(\mathbf{x}_t, t) \rVert ^ 2 \Bigr]
\end{equation}

Unfortunately, due to these sequential evaluations, diffusion models are computationally expensive. To address this, latent diffusion models (LDMs) \cite{stable-diffusion} first employ a VQ-VAE \cite{vq-vae} to transform the high-dimensional image space into a lower-dimensional latent space. This significantly reduces computational complexity and allows the diffusion model to focus on semantics rather than the imperceptible details. Formally, the encoder function $\mathcal{E}$ of the VQ-VAE takes an input image $\mathbf{x} \in \mathbb{R}^{H \times W \times 3}$ and generates a corresponding latent representation $\mathbf{z} \in \mathbb{R}^{h \times w \times c}$ with $h < H$ and $w < W$. These latents $\mathbf{z}$ are then used to train the diffusion model. The loss function of the diffusion model has to be modified accordingly to account for the use of the latent representation (see Equation \ref{eq:loss-function-ldm}). The predicted output of a trained diffusion model can be translated back into image space via the use of the decoder from the VQ-VAE.

\begin{equation}
    \label{eq:loss-function-ldm}
    L_{LDM} := \mathbb{E}_{\mathcal{E}(\mathbf{x}), \epsilon \sim \mathcal{N}(\mathbf{0}, \mathbf{I}), t} \Bigr[ \lVert \epsilon - \epsilon_\theta(\mathbf{z}_t, t) \rVert ^ 2 \Bigr]
\end{equation}

\subsubsection{Conditioning Mechanisms}

These image synthesis diffusion models are also capable of modelling conditional distributions $q(\mathbf{x} | y)$ by implementing $p_\theta(\mathbf{x}_{t-1} | \mathbf{x}_{t}, y)$ with a conditional denoising autoencoder, which is denoted as $\epsilon_\theta(\mathbf{z}_t, t, y)$. Recent studies have shown that $y$ can be obtained from various modalities such as class labels, text prompts or other images \cite{stable-diffusion, diffusion-beats-gan}.

The latent diffusion study \cite{stable-diffusion} implemented $\epsilon_\theta(\mathbf{z}_t, t, y)$ by adding cross-attention modules \cite{transformers} into intermediate layers of the UNet architecture. This was based on prior research that showed that cross-attention can be applied to a wide variety of modalities besides text \cite{perceiver-io,perceiver}. While in the transformer paper cross-attention is performed on the output sentence conditioned on the input sentence, in this context, cross-attention is performed on flattened intermediate feature maps of the UNet conditioned on a vectorized representation of $y$. To obtain a vector representation of $y$, a domain-specific encoder $\tau_\theta$ is introduced. For text conditional synthesis, $\tau_\theta$ is implemented as an unmasked transformer that processes a BERT \cite{BERT} encoded version of the input text prompt $y$.

The conditional model is trained based on image and condition pairs via Equation \ref{eq:loss-function-conditional-ldm}, where both $\tau_\theta$ and $\epsilon_\theta$ are jointly optimised. It is important to note that in Stable Diffusion \cite{github-stable-diffusion}, the unmasked transformer was replaced with a frozen version of CLIP \cite{clip}, thus only requiring the optimisation of $\epsilon_\theta$.

\begin{equation}
    \label{eq:loss-function-conditional-ldm}
    L_{LDM} := \mathbb{E}_{\mathcal{E}(\mathbf{x}), \epsilon \sim \mathcal{N}(\mathbf{0}, \mathbf{I}), t} \Bigr[ \lVert \epsilon - \epsilon_\theta(\mathbf{z}_t, t, \tau_\theta(y)) \rVert ^ 2 \Bigr]
\end{equation}

\subsubsection{Fine-tuning}

The DreamBooth technique \cite{ruiz2022dreambooth}, showcases the potential of fine-tuning a pretrained text conditional diffusion model to generate new renditions of a given object by associating it with specific pseudo-words. In this approach, a limited set of images $\mathcal{X} = \{x_i | i \in {1, ..., N}\}$ depicting the same object, all with the same conditioning text $y_s$, are used to fine-tune the model. This results in the model learning to associate $y_s$ with the object illustrated in $\mathcal{X}$

The authors propose a very simple format for the condition text $y_s$, as illustrated in Equation \ref{eq:cond-vec-pseudo}. Here, the [identifier] refers to the pseudo-word associated with the subject, while the [class noun] serves as a coarse class descriptor of the subject (e.g., dog, apple, tree, etc.). It is important to note that not all words can serve as suitable pseudo-words, and the authors suggest identifying relatively rare tokens in the vocabulary of the word tokenizer \footnote{In our implementation, we adopt the use of the pseudo-word 'sks' as proposed by the \textit{diffusers} implementation \cite{github-diffusers}.}.

\begin{align}
    \label{eq:cond-vec-pseudo}
    y_s := \text{"a [identifier] [class noun]"} && \text{E.g. } y_s = \text{"a sks tree"}
\end{align}

With the pairs of $(x_i, y_s)$ obtained in this manner, the pretrained model could resume training by using the standard conditional $L_{LDM}$ loss. However, to avoid overfitting and language drift the authors propose a so-called class-specific prior preservation loss, which is a technique that involves regularising the model with its own generated samples. Specifically, the authors generate a second set of images $\mathcal{X}^{(pr)}$ using a frozen version of the pretrained model prior to any fine-tuning. These images are generated using a conditioning text $y^{(pr)}_s$ that is equivalent to $y_s$ but omits the pseudo-word as shown in Equation \ref{eq:cond-vec}. The purpose of this regularisation is to encourage the model to produce output images that are similar to those of the frozen network when the pseudo-word is not present, thereby retaining its prior knowledge of $\text{[class noun]}$.

\begin{align}
    \label{eq:cond-vec}
    y^{(pr)}_s := \text{"a [class noun]"} && \text{E.g. } y^{(pr)}_s = \text{"a tree"}
\end{align}

Given the two sets $\mathcal{X}$ and $\mathcal{X}^{(pr)}$, the loss function is defined as in Equation \ref{eq:loss-function-dreambooth}. Here, $\mathbf{x}$ and $\mathbf{x}^{(pr)}$ are drawn from the sets $\mathcal{X}$ and $\mathcal{X}^{(pr)}$, respectively. The variables $\epsilon$ and $\epsilon'$ are sampled from the normal distribution $\mathcal{N}(\mathbf{0}, \mathbf{I})$, and $\lambda$ controls the weight of the regularisation. While DreamBooth focuses on Imagen \cite{imagen}, the described loss function has been reformulated to be compatible with latent diffusion models. See Appendix \ref{appendix:b} for the implementation details.

\begin{multline}
    \label{eq:loss-function-dreambooth}
    L_{DreamBooth} := \mathbb{E}_{\mathcal{E}(\mathbf{x}), \mathcal{E}(\mathbf{x}^{(pr)}), \epsilon, \epsilon', t} \Bigr[ \lVert \epsilon - \epsilon_\theta(\mathbf{z}_t, t, \tau_\theta(y_s)) \rVert ^ 2 \\ 
    + \lambda \lVert \epsilon' - \epsilon_\theta(\mathbf{z}^{(pr)}_t, t, \tau_\theta(y^{(pr)})) \rVert ^ 2 \Bigr]
\end{multline}

\subsection{Genfusion: Image Generator}

\begin{figure*}[!htb]
    \centering
    \begin{minipage}{1\textwidth}
	    \centering
        \subfloat[\textbf{Generated Trees with Green Apples} \label{fig:genfusion-green}]{\includegraphics[width=0.95\textwidth]{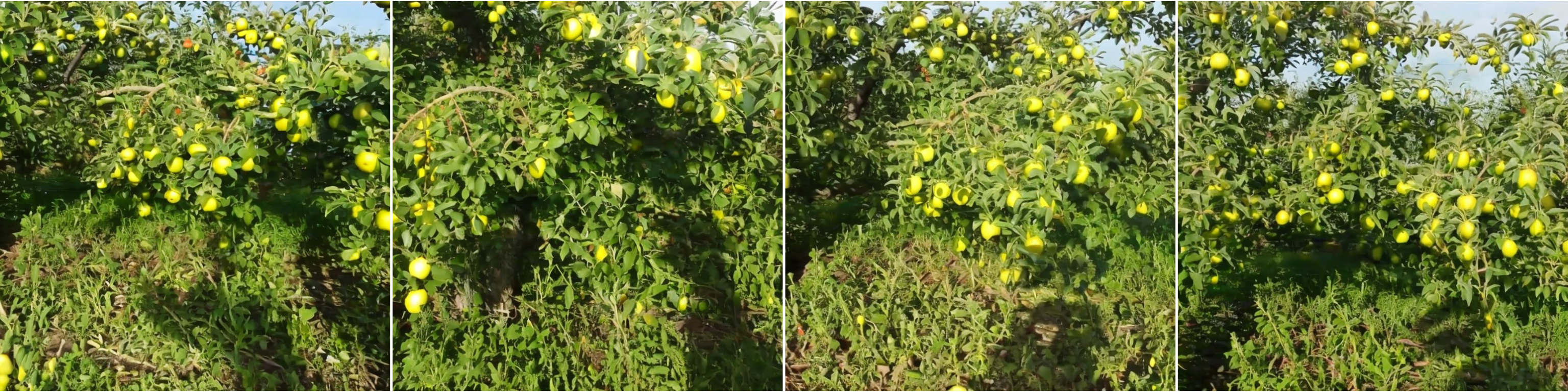}}
	\end{minipage}%
	\hspace{0.01\textwidth}
	\begin{minipage}{1\textwidth}
	    \centering
        \subfloat[\textbf{Generated Trees with Red Apples} \label{fig:genfusion-red}]{\includegraphics[width=0.95\textwidth]{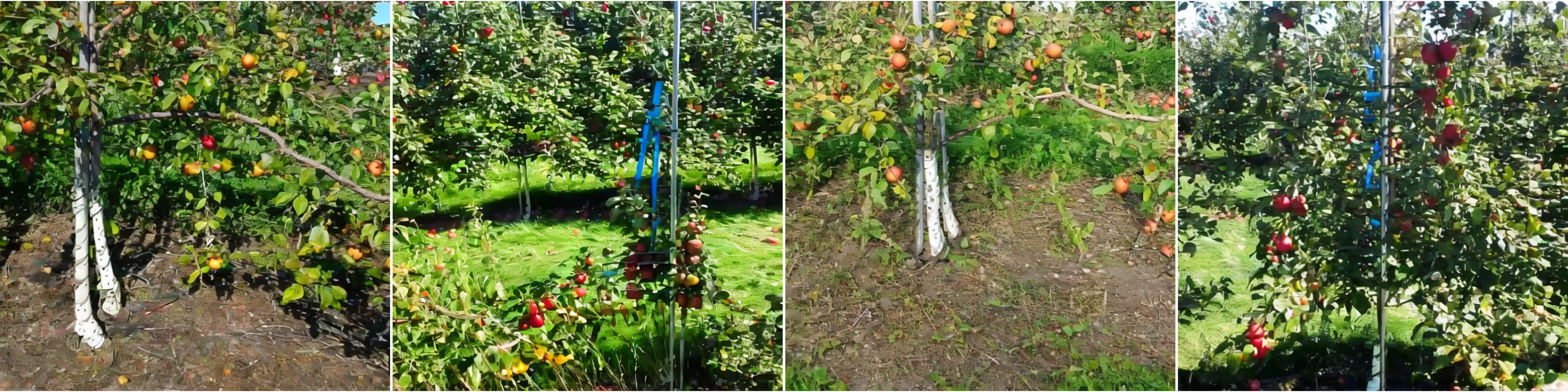}}
	\end{minipage}
    \caption{A subset of the generated images from our Genfusion image generation pipeline. For additional images, please refer to Appendix \ref{appendix:c}.}
    \label{fig:images-generated}
\end{figure*}

Our objective was to generate an artificial training set by selecting a small subset of images from the MinneApple training set for fine-tuning purposes. Our study was based on the Stable Diffusion v2.1 model as described in \cite{huggingface-stable-diffusion-version-2}. However, one of the major challenges we encountered was to determine an optimal number of images that would capture the diversity present in MinneApple.

The DreamBooth authors \cite{ruiz2022dreambooth} state that it is possible to fine-tune a diffusion model with a mere 3 to 5 images of the subject to replicate. Nonetheless, our findings indicate that to capture the diversity present in the MinneApple dataset, we required no less than 20 images. We aimed to preserve the same distribution of apple colours, namely, 54 green and 482 red. To achieve this distribution in our generated dataset, we fine-tuned the model twice. The first fine-tuning was done using 10 green apple tree images, while the second was done using 10 red apple tree images. These 20 images were carefully selected from the MinneApple train set, with an emphasis on optimising for a diverse range of samples. During the generation process, we used the green apple tree model 54 times and the red apple tree model 482 times to maintain the original distribution of apple colours in the generated dataset. The complete dataset generation pipeline, hereafter referred to as Genfusion, is visually represented in Figure \ref{fig:genfusion}.

In relation to the prompts employed during the fine-tuning process, we defined $y_s$ and $y_s^{(pr)}$ to be equivalent to the example prompts outlined in Equations \ref{eq:cond-vec-pseudo} and \ref{eq:cond-vec}, respectively. Regarding the image data, we defined the set $\mathcal{X}$ to be either 10 green apple trees or 10 red apple trees sampled from MinneApple. Furthermore, the prior preservation dataset $\mathcal{X}^{(pr)}$ consisted of 200 images that were generated by the model before the fine-tuning process. See Figure \ref{fig:images-generated} for a subset of our generated images \footnote{All images in $\mathcal{X}$ and a subset of $\mathcal{X}^{(pr)}$ are shown in Appendix \ref{appendix:d}.}.

\subsubsection{Annotations}

To reduce the effort required for annotation, we employed the baseline model's predictions on the MinneApple training set as a preliminary proposal for bounding box annotations on our generated data. We then proceeded to manually refine these predictions by eliminating false positives and identifying apples that the detector had missed. It is important to note that our approach is not constrained by the requirement for a baseline model, as annotations could also have been carried out entirely through manual means.

\subsubsection{Preliminary Analysis: Generating Annotations}

We did an investigation into how this fine-tuning methodology could generalise to also automatically generate annotations. This section provides a brief overview of the ideas that were explored. However, due to current limitations, the results were not included in the main experiment section. More details can be found in Appendix \ref{appendix:e}.

Given the exceptional performance of diffusion models in generating image data, the possibility of encoding annotation data within the image was explored. Specifically, adding a new fourth channel encoding locations with high-intensity dots at the centre of objects. This method proved advantageous in scenarios with overlapping objects, as the dots were easier to extract compared to overlapping outlines of bounding boxes. However, it should be noted that this encoding method does not preserve the width and height information of the objects. Fortunately, recent works, such as Segment Anything \cite{sam}, have shown promise by using dots as guides to create masks that accurately represent the object's width and height in the corresponding image location.

Unfortunately, the addition of a fourth channel to the input proved to be problematic since the VQ-VAE from latent diffusion was incapable of processing inputs with more than three channels. To overcome this limitation, we conducted a preliminary analysis by repurposing the blue channel as the annotation channel. Although this decision may be regarded as somewhat dubious, it would adequately serve the purpose of conducting an initial analysis on the capabilities of this DreamBooth fine-tuning. Nonetheless, to ensure scientific validity in our comparisons, it is necessary to contrast it with a detector that has been exclusively trained on the red and green information of the MinneApple dataset.

Upon fine-tuning the model on images in which the first two channels consisted of colour information and the third channel held our encoded annotations, we noted that the model seemed to be capable of generating images with accurate annotations. The primary limitation of this approach was that the information contained in the annotation channel seemed to "leak" to the other channels, resulting in high-intensity dots at the centre of the apples in the red and green channels. As expected, this resulted in severe overfitting when training the model on the green and red colour channels.

An alternative option was to encode annotations as instance segmentation masks. However, this method had limitations. Retraining the VQ-VAE from latent diffusion would be necessary to preserve small pixel intensity differences after encoding and decoding. Unfortunately, within the scope of our study, undertaking such a retraining process was not feasible.

\subsection{YOLO Object Detectors}

YOLOv5 and YOLOv8 are two architectures from the YOLO (You Only Look Once) family \cite{yolov1,redmon2017yolo9000,redmon2018yolov3}. YOLOv5 \cite{yolov5} utilises anchor boxes to predict the bounding boxes and employs a three-stage approach, which includes a backbone network, a neck network, and a head network. The backbone network is based on the CSPDarknet architecture \cite{wang2020cspnet}, which incorporates depthwise convolutions to improve computational efficiency. The neck network is a Path Aggregation Network (PANet) \cite{liu2018path} that merges the features from the backbone network to produce a feature map with high semantic information. The head network then performs detection by predicting the bounding boxes and class probabilities from the feature map.

In contrast, YOLOv8 \cite{yolov8} lacks a published paper regarding its design as of the current writing. However, an examination of the code repository reveals that YOLOv8 employs an anchor-free approach, reducing box predictions and accelerating the Non-Maximum Suppression (NMS) process. Both YOLOv5 and YOLOv8 repositories offer default image augmentations, but for a fair comparison, all augmentations were disabled. Additionally, both architectures provide five scaled versions (nano, small, medium, large, and extra-large), and this study presents results for all sizes.

\subsection{Evaluation Metrics}

To ensure a comprehensive evaluation of the object detectors, we will utilise established evaluation metrics that have been employed in other object detection datasets \cite{hani2020minneapple,everingham2009pascal,lin2014microsoft}. Specifically, we will report the Average Precision (AP) \cite{everingham2009pascal} to summarise the precision-recall curve.

Formally, predicted bounding boxes are assigned to ground truth bounding boxes by calculating their IoU (see Equation \ref{eq:iou}). To be considered a correct detection, the IoU between the predicted bounding box $B_p$ and ground truth bounding box $B_{gt}$ needs to exceed a predefined threshold. We present, in line with other studies \cite{everingham2009pascal,lin2014microsoft}, our results with a threshold of 0.5 and 0.75.

\begin{equation}
\label{eq:iou}
\text{IoU} := \frac{B_p \cap B_{gt}}{B_p \cup B_{gt}}
\end{equation}

\begin{figure*}[!htb]
    \centering
    \newcolumntype{C}[1]{>{\centering\arraybackslash}p{#1}}
    \begin{tabular}{@{}C{0.3\textwidth}C{0.333\textwidth}C{0.3\textwidth}@{}}
        Tree 1 & Tree 2 & Tree 3
    \end{tabular}
    \begin{minipage}{1\textwidth}
	    \centering
        \subfloat[\textbf{MinneApple Baseline (YOLOv8l)} \label{fig:qualitative-baseline}]{\includegraphics[width=1\textwidth]{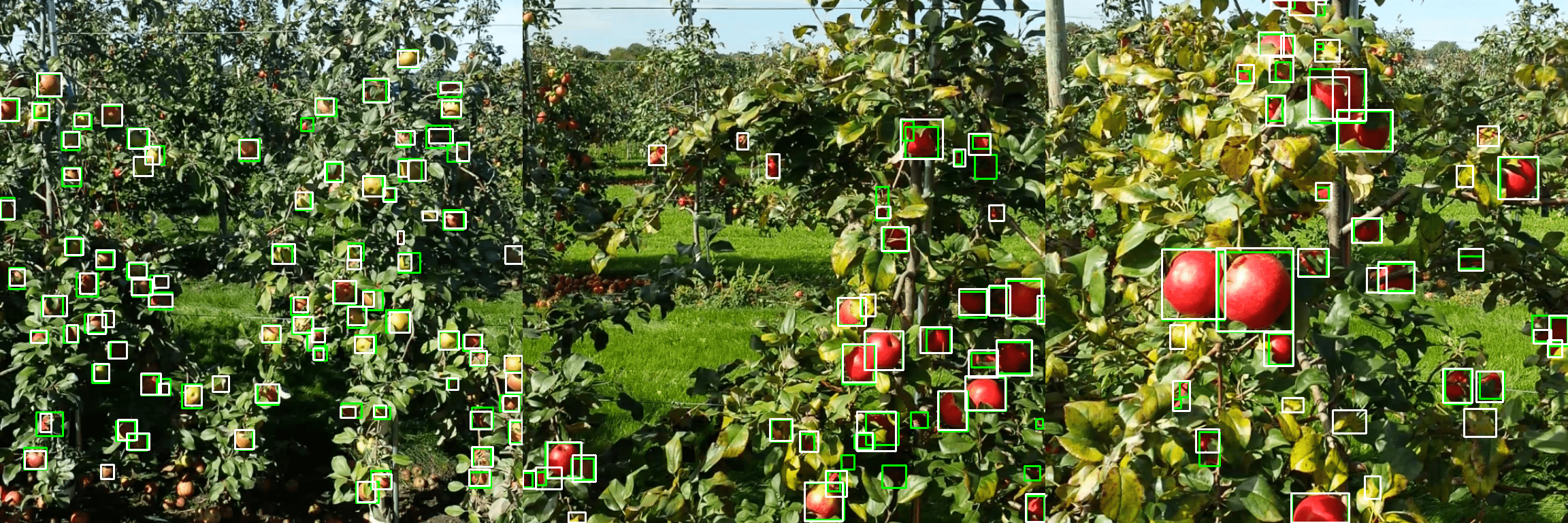}}
	\end{minipage}%
	\hspace{0.01\textwidth}
	\begin{minipage}{1\textwidth}
	    \centering
        \subfloat[\textbf{Genfusion (YOLOv8l)} \label{fig:qualitative-genfusion}]{\includegraphics[width=1\textwidth]{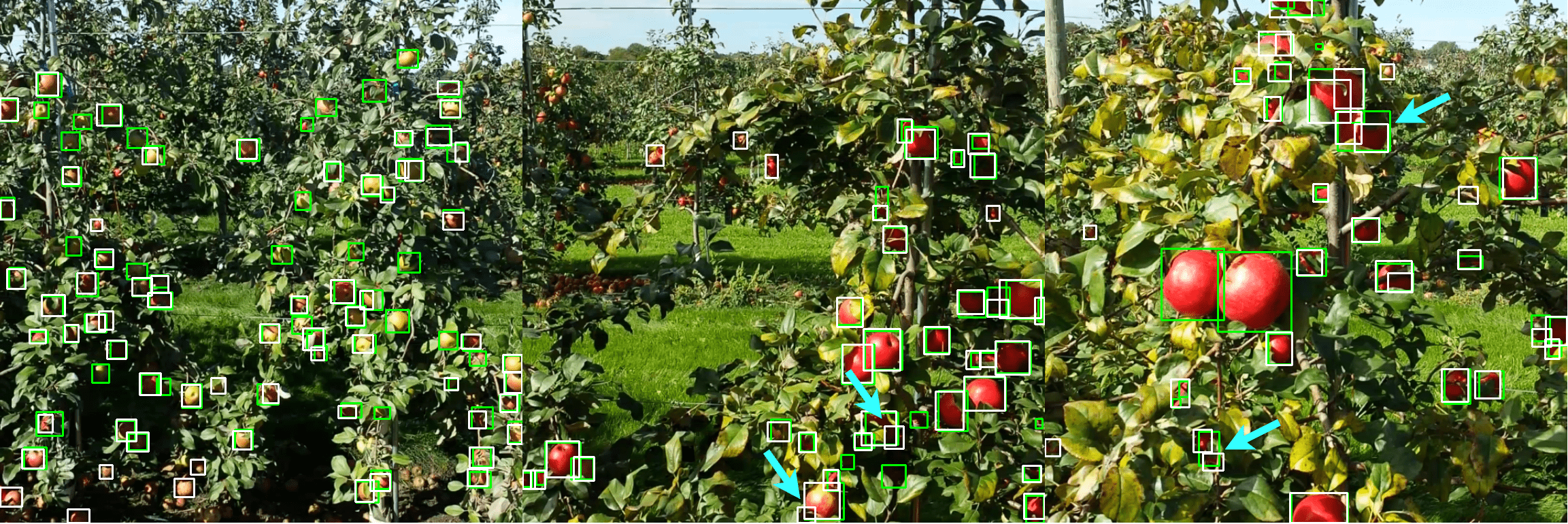}}
	\end{minipage}
    \caption{The predicted (white) and ground truth (green) bounding boxes of the YOLOv8l Baseline and Genfusion model from three of the images from the MinneApple test set. The blue arrows indicate instances where the Genfusion model predicts two bounding boxes for a single apple.}
    \label{fig:qualitative-results}
\end{figure*}

After assigning predictions to ground truths, precision and recall are calculated at multiple confidence thresholds, resulting in a precision-recall curve denoted as $p(r)$. The average precision (AP) is computed as the mean precision at specific recall levels ($\mathcal{R} = {0.0, 0.01, ..., 1}$). To mitigate fluctuations in $p(r)$, the authors \cite{everingham2009pascal} proposed interpolating precision by selecting the maximum precision for recall levels greater than or equal to $r$ (Equation \ref{eq:ap}).

\begin{align}
    \label{eq:ap}
    \text{AP} := \frac{1}{|\mathcal{R}|}\sum_{r \in \mathcal{R}} p_{interp}(r), && p_{interp}(r) := \max_{\tilde{r}: \tilde{r} \geq r} p(\tilde{r})
\end{align}

In addition to reporting the evaluation metrics at IoU thresholds of 0.5 (AP@0.5) and 0.75 (AP@0.75). We use the COCO challenge evaluation metric \cite{lin2014microsoft} as our primary performance measure, which calculates the average AP scores over ten IoU thresholds ranging from 0.5 to 0.95 in increments of 0.05 (AP@0.5:0.05:0.95).

\section{Results}

\begin{figure}[H]
    \centering
    \includegraphics[width=0.42\textwidth]{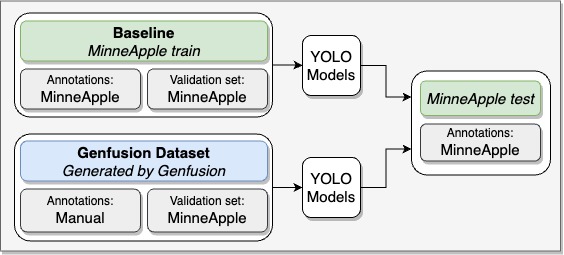}
    \caption{Experiment Setup}
    \label{fig:experiment-1}
\end{figure}

The methodology proposed in this study has been specifically designed to evaluate the effectiveness of fine-tuning text-to-image diffusion models for the purpose of generating datasets. The objective was to create images that closely resemble the MinneApple benchmark dataset \cite{hani2020minneapple}. To establish the effectiveness of our approach, we manually annotated the generated images and employed them to train various YOLOv5 and YOLOv8 models. Our experimental setup is illustrated in Figure \ref{fig:experiment-1}. Furthermore, a preliminary analysis of an automated annotation generation technique is presented in Appendix \ref{appendix:e} but was excluded from this section due to unsatisfactory results.

Our experimental results, which are presented in Table \ref{tab:ap-metrics}, compare the performance of models trained exclusively on the MinneApple train set (baseline) with those trained on our generated data (Genfusion). Results are presented for a wide range of YOLOv5 \cite{yolov5} and YOLOv8 \cite{yolov8} model sizes. To account for uncertainties, we trained each model five times, and the table reports the mean and standard deviation of the AP metrics.

\newcommand{\experiment}[2]{\renewcommand{\arraystretch}{0.5}\begin{tabular}{@{}r@{}}\renewcommand{\arraystretch}{1}#1 \\ \small{#2}\end{tabular}}
\newcommand{\multirows}[2]{\renewcommand{\arraystretch}{0.5}\begin{tabular}{@{}l@{}}\renewcommand{\arraystretch}{1}#1 \\ #2\end{tabular}}
\newcommand{\substract}[1]{\multirows{\rule{25pt}{0.7pt} \rule{5pt}{0.2pt}}{\textbf{#1}}}

\begin{table}[H]
\centering
\caption{AP evaluation metrics on the MinneApple test set, using different YOLOv5 and YOLOv8 models trained over the generated train set, compared to the model trained over the MinneApple train set.}
\renewcommand{\arraystretch}{2}
\begin{tabular}{
    r|lll
}
\hline
\textbf{Dataset} & \textbf{AP@0.5:0.05:0.95} & \textbf{AP@0.50} & \textbf{AP@0.75} \\    \hline
\hline

\experiment{Baseline}{yolov8x} & 0.45 \small{±0.005} & 0.79 \small{±0.005} & 0.47 \small{±0.011} \\
\experiment{Genfusion}{yolov8x} & 0.33 \small{±0.050} & 0.66 \small{±0.060} & 0.29 \small{±0.070} \\[-7pt]
& \substract{0.12} & \substract{0.13} & \substract{0.18} \\\hline

\experiment{Baseline}{yolov8l} & 0.45 \small{±0.008} & 0.78 \small{±0.009} & 0.47 \small{±0.008} \\
\experiment{Genfusion}{yolov8l} & 0.35 \small{±0.019} & 0.70 \small{±0.011} & 0.32 \small{±0.031} \\[-7pt]
& \substract{0.10} & \substract{0.08} & \substract{0.15} \\\hline

\experiment{Baseline}{yolov8m} & 0.44 \small{±0.008} & 0.77 \small{±0.009} & 0.46 \small{±0.010} \\
\experiment{Genfusion}{yolov8m} & 0.33 \small{±0.008} & 0.66 \small{±0.016} & 0.30 \small{±0.013} \\[-7pt]
& \substract{0.11} & \substract{0.11} & \substract{0.16} \\\hline

\experiment{Baseline}{yolov8s} & 0.43 \small{±0.009} & 0.75 \small{±0.005} & 0.44 \small{±0.009} \\
\experiment{Genfusion}{yolov8s} & 0.32 \small{±0.024} & 0.66 \small{±0.024} & 0.28 \small{±0.036} \\[-7pt]
& \substract{0.11} & \substract{0.09} & \substract{0.16} \\\hline

\experiment{Baseline}{yolov8n} & 0.40 \small{±0.005} & 0.73 \small{±0.004} & 0.39 \small{±0.005} \\
\experiment{Genfusion}{yolov8n} & 0.31 \small{±0.012} & 0.66 \small{±0.015} & 0.27 \small{±0.012} \\[-7pt]
& \substract{0.09} & \substract{0.07} & \substract{0.12} \\\hline\hline

\experiment{Baseline}{yolov5x} & 0.45 \small{±0.010} & 0.77 \small{±0.013} & 0.47 \small{±0.013} \\
\experiment{Genfusion}{yolov5x} & 0.35 \small{±0.011} & 0.69 \small{±0.011} & 0.31 \small{±0.023} \\[-7pt]
& \substract{0.10} & \substract{0.08} & \substract{0.16} \\\hline

\experiment{Baseline}{yolov5l} & 0.44 \small{±0.008} & 0.77 \small{±0.010} & 0.46 \small{±0.011} \\
\experiment{Genfusion}{yolov5l} & 0.33 \small{±0.028} & 0.67 \small{±0.031} & 0.30 \small{±0.038} \\[-7pt]
& \substract{0.11} & \substract{0.10} & \substract{0.16} \\\hline

\experiment{Baseline}{yolov5m} & 0.43 \small{±0.007} & 0.76 \small{±0.005} & 0.44 \small{±0.009} \\
\experiment{Genfusion}{yolov5m} & 0.33 \small{±0.016} & 0.67 \small{±0.016} & 0.29 \small{±0.022} \\[-7pt]
& \substract{0.10} & \substract{0.09} & \substract{0.15} \\\hline

\experiment{Baseline}{yolov5s} & 0.42 \small{±0.009} & 0.75 \small{±0.009} & 0.43 \small{±0.010} \\
\experiment{Genfusion}{yolov5s} & 0.32 \small{±0.016} & 0.66 \small{±0.023} & 0.28 \small{±0.031} \\[-7pt]
& \substract{0.10} & \substract{0.09} & \substract{0.15} \\\hline

\experiment{Baseline}{yolov5n} & 0.39 \small{±0.004} & 0.73 \small{±0.005} & 0.39 \small{±0.009} \\
\experiment{Genfusion}{yolov5n} & 0.27 \small{±0.028} & 0.61 \small{±0.043} & 0.20 \small{±0.037} \\[-7pt]
& \substract{0.12} & \substract{0.12} & \substract{0.19} \\\hline

\hline
\end{tabular}%
\label{tab:ap-metrics}%
\end{table}

As can be seen, the baseline models achieve the highest AP scores. However, our generated data demonstrated relatively good performance in comparison. The AP difference between the baseline and our method is consistently between 0.09 and 0.12 for AP@0.5:0.05:0.95. However, for AP@0.75, the difference is higher, ranging from 0.12 to 0.19, indicating that with our method the locations of the boxes are less precise. Additionally, as expected, our findings confirm that larger models outperform smaller ones with the highest AP of 0.35 for Genfusion using the YOLOv8l and YOLOv5x models. Interestingly, YOLOv5 and YOLOv8 exhibit similar AP scores, suggesting that the improvements in YOLOv8 do not significantly affect detection performance for this task.

To comprehend the implications of a 0.10 AP difference in terms of detections, we conducted a qualitative analysis on predictions of the YOLOv8l model. Our evaluation revealed that the baseline and generated approaches produced similar outcomes, as illustrated in Figure \ref{fig:qualitative-results}. The detectors demonstrated a reasonable ability to predict the location of apples in the trees. As expected, in cases where the baseline failed to detect a specific apple, the generated approach also showed similar failure rates, particularly for instances located in areas with high levels of shading or occlusions.

Notably, when comparing the Genfusion model with the baseline, the former primarily encountered difficulty in detecting very small apples (Tree 1) or very large apples (Tree 3). In addition, the results show that the Genfusion model had a slight tendency to detect a single apple as two separate entities in cases where they were large or shaded, as indicated by the blue arrows, contributing to its lower AP scores at 0.75 IoU.

\section{Discussion}

Although the present study shows diffusion models in image generation as a promising research direction, our findings reveal a limitation inherent in either our data generation or annotation process. Specifically, our detectors exhibit difficulty detecting apples that are exceptionally large or small. When looking at the generated data it became apparent that our data generator was incapable of mimicking these scenarios. This could be because they were not subjected to a diverse enough representation of the real-world scenario. In addition, it is possible that our manual annotation quality falls short of the standards set by MinneApple, who reported spending up to 30 minutes annotating a single image \cite{hani2020minneapple}; including in-person reviews to ensure annotation consistency.

With regards to the automatic annotation generation approach described in Appendix \ref{appendix:e}, its complete potential is yet to be fully explored. However, the findings obtained in this study serve to showcase the versatility of this fine-tuning approach and suggest promising avenues for future research. Specifically, future work could investigate the origin of the observed channel leakage and explore alternative loss functions that may help regulate information flow from the annotation channel to other channels.

Our investigation only demonstrated the feasibility of this approach for apple detection purposes. Nevertheless, we hypothesise that this methodology has the potential to be extended to other domains based on the capability of pretrained diffusion models to effectively model diverse image distributions. Ultimately, our research highlights the potential for this methodology to be applied successfully in real-world scenarios that are characterised by a scarcity of data, particularly in domains where data collection is challenging.

\section{Conclusion}

In conclusion, our proposed framework for generating training images for apple detection in orchards exhibits promising results and addresses two key research questions. Firstly, we demonstrate that a pretrained diffusion model can be effectively fine-tuned using the Dreambooth technique to generate a dataset representative of a specific real-world scenario. This process enables the generation of high-quality synthetic images, bridging the gap between limited real data availability and the need for extensive datasets. As a result, it opens new possibilities for employing deep learning techniques in data scares domains.

Secondly, we investigate how the performance of an object detector trained on data generated with our framework, named Genfusion, compares to one trained on real data when evaluated in a real-world scenario. Although our approach does not achieve complete parity with the performance of models trained on real-world data, it yields remarkably close approximations, with an average precision (AP) difference ranging from 0.09 to 0.12. We hypothesise that when fine-tuning with a larger and more diverse set of orchard samples, our approach could rival the performance of real-world data. However, such a trade-off necessitates careful consideration of the time taken for data acquisition and the performance achieved.

\section{Future Work}

Further research endeavours can explore the refinement of the fine-tuning process by employing a larger or more diverse collection of images, aiming to approach the performance level achieved by real-world data. The results presented in this paper also warrant additional investigation into text-to-image dataset generation for more complex detection tasks. For instance, replication of renowned object detection benchmark datasets such as COCO \cite{lin2014microsoft} or Pascal VOC \cite{everingham2009pascal} can provide valuable insights into the robustness and scalability of the proposed framework.

Moreover, it is advisable for future work to explore alternative prompting techniques. In this study, we focused on training and inference utilising the prompt "A sks tree". However, the versatility of diffusion models would enable their utilisation for augmentations involving varied prompts, such as "A sks tree in the shadow" or "A sks tree in the snow". Exploring these alternative prompting strategies can potentially enhance the model's ability to capture and interpret diverse contextual information, leading to improved performance and adaptability in real-world scenarios.

\section*{Broader Impact}

The present study expands upon the pre-existing body of literature regarding diffusion models, DreamBooth fine-tuning, and deep object detectors by integrating them into one coherent framework. This research is a notable advancement in simplifying the data acquisition process in vision based domains.

Unfortunately, generative models have a significant downside as they can be employed for malevolent purposes. One of the most concerning applications is the generation of fake images and videos of prominent figures for political motives. These fabricated images are a contemporary problem and have become significantly more widespread with the advent of generative models. Moreover, the easy training of classifiers or detection models in combination with these fabricated images presents a complex issue, amplifying the potential for misuse.

On the other hand, Genfusion and comparable data generation methodologies can offer significant advantages in improving object detection and recognition in data-sparse domains. In the agricultural industry, for example, limited availability of data relating to specific outliers, such as particular diseases, can make it challenging to develop accurate and effective crop disease management tools. In addition, medical imaging datasets often face limitations owing to privacy concerns, posing a challenge for training computer vision models to perform image analysis for medical assistance. By generating supplementary data to train computer vision models, such techniques can potentially enhance the accuracy and reliability of such systems, thus leading to substantial practical benefits.

In conclusion, the use of generated image data for training deep models presents both opportunities and ethical challenges. While these approaches can enhance the accuracy and reliability of computer vision systems and their practical benefits, it is essential to address the potential malicious applications. Therefore, it is essential to continue exploring and implementing ethical practices that balance the benefits and risks of these techniques. By doing so, we can ensure that these advancements in this research area contribute positively to the broader world.

\acknowledgements{
    \begin{itemize}
        \item This project is financially supported by Regieorgaan SIA (part of NWO) and performed within the RAAK PRO project Mars4Earth.
        \item We would like to thank our collaborators at Saxion University of Applied Sciences for insightful discussions.
    \end{itemize}
    \begin{figure}[!htb]
        \includegraphics[width=1.6in]{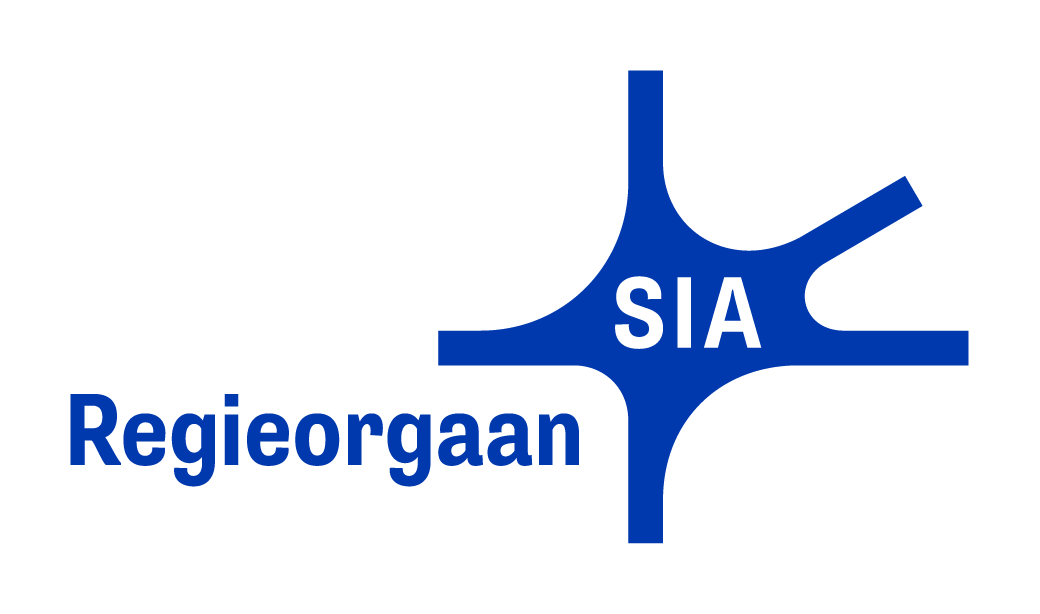}
        \hfill
        \includegraphics[width=1.6in]{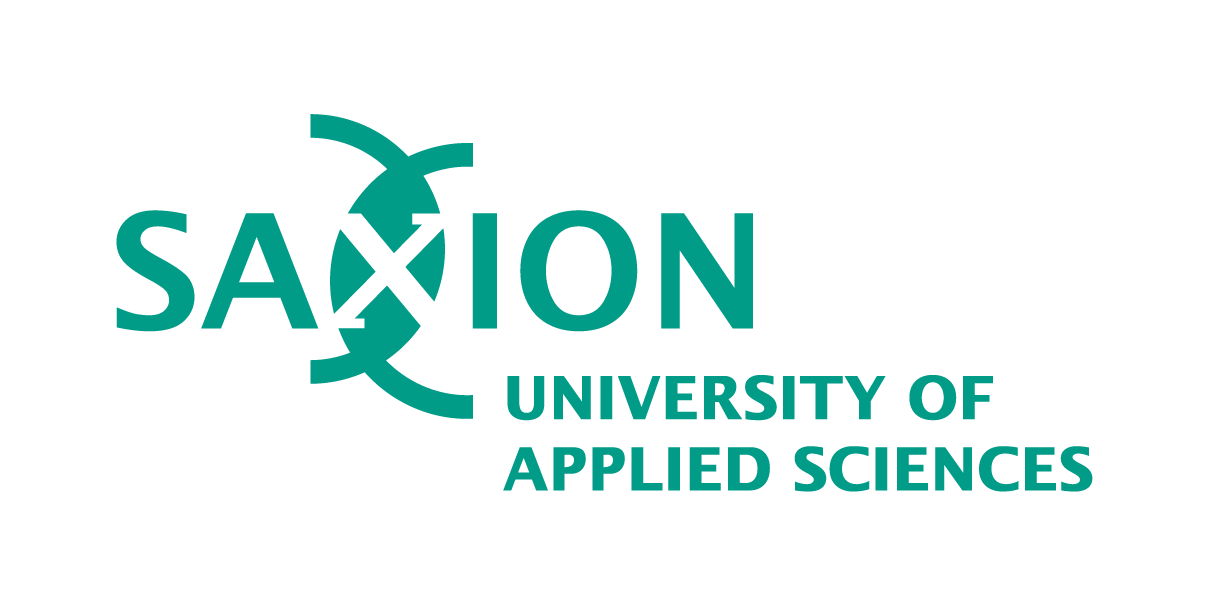}
    \end{figure}
}

\bibliographystyle{unsrt}
\bibliography{sources}

\clearpage
\begin{appendices}
    \onecolumn
	\section{Time-conditional UNet}
    \label{appendix:a}

\begin{figure}[H]
        \centering
        \includegraphics[width=0.83\textwidth]{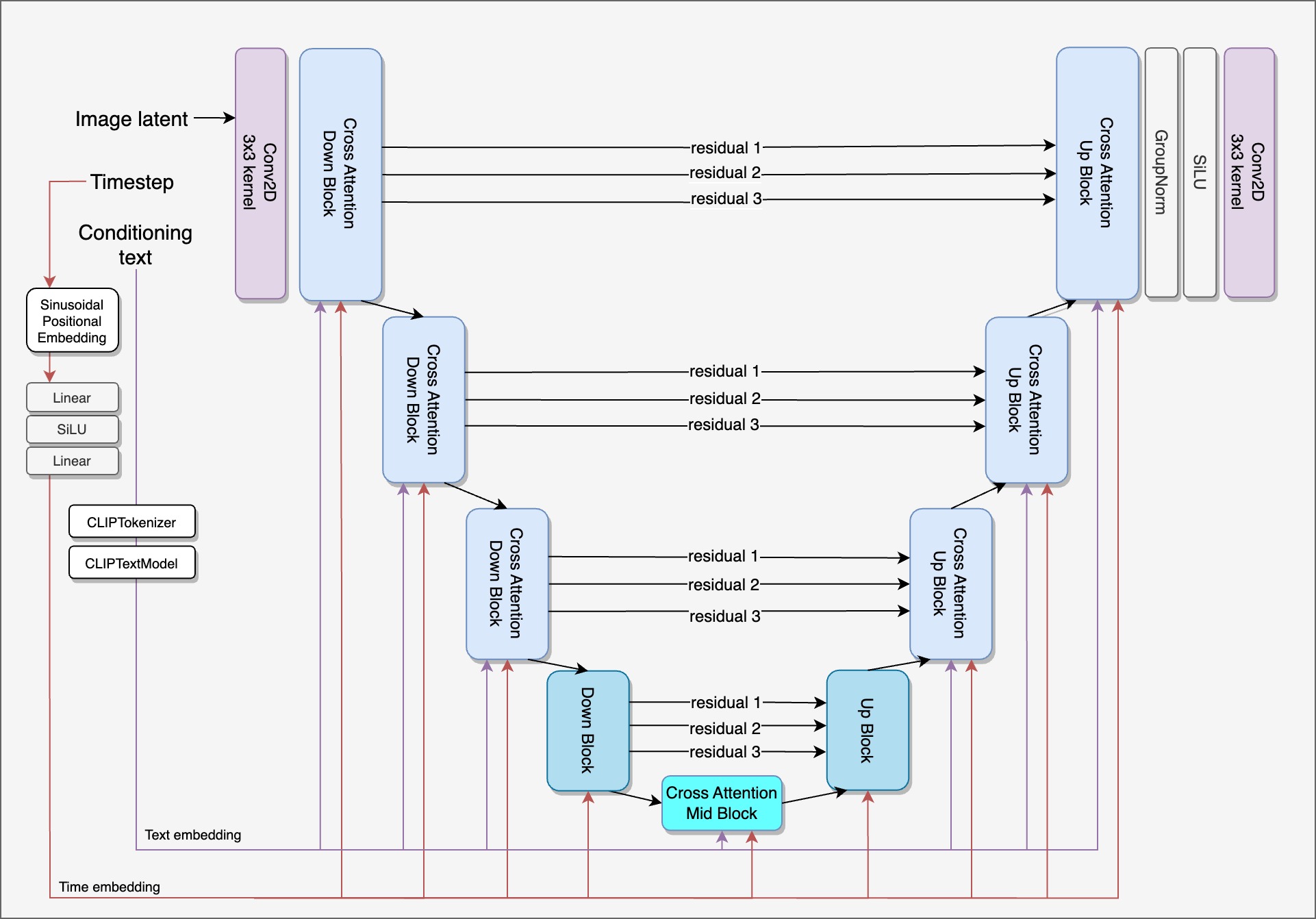}
        \caption{The time-conditional UNet architecture}
        \label{fig:unet-appendix}
    \end{figure}

    This research uses Stable Diffusion v2.1 \cite{huggingface-stable-diffusion-version-2} published by Stability AI. This version is trained at a resolution of 768x768. It has the same number of parameters in the UNet as version 1.5 but uses OpenCLIP-ViT/H \cite{open-clip} as the text encoder. Interestingly, it was observed that comprehensive diagrams illustrating the UNet architecture were unavailable. Consequently, we took the initiative to create these diagrams based on an in-depth analysis of the code. A high-level overview of the architecture can be found in Figure \ref{fig:unet-appendix}.

\begin{figure}[H]
        \centering
        \includegraphics[width=0.76\textwidth]{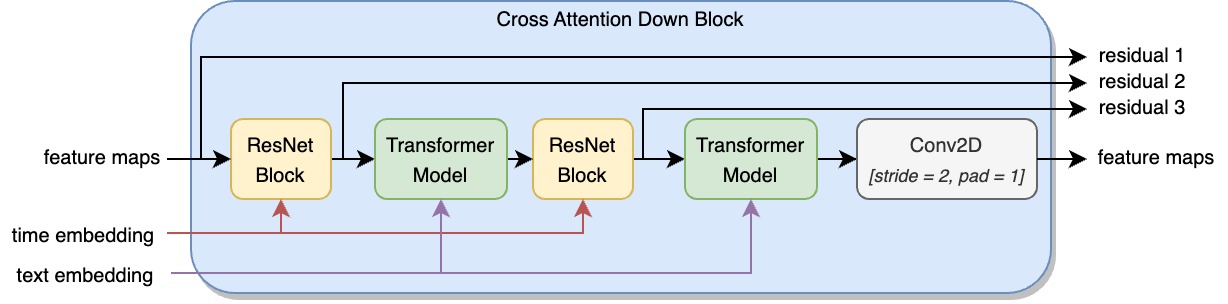}
        \caption{Cross Attention Down Block module from the time-conditional UNet architecture}
        \label{fig:cross-attention-down-block}
    \end{figure}

\begin{figure}[H]
        \centering
        \includegraphics[width=0.76\textwidth]{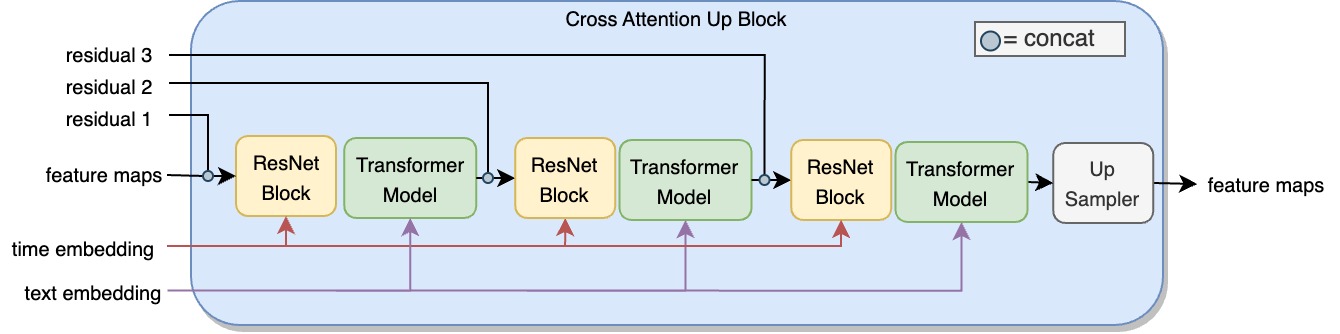}
        \caption{Cross Attention Up Block module from the time-conditional UNet architecture}
        \label{fig:cross-attention-up-block}
    \end{figure}

\begin{figure}[H]
        \centering
    
    	\begin{minipage}{0.49\textwidth}
    	    \centering
            \subfloat[\textbf{Up Block (without attention)} \label{fig:up-block}]{\includegraphics[width=1\textwidth]{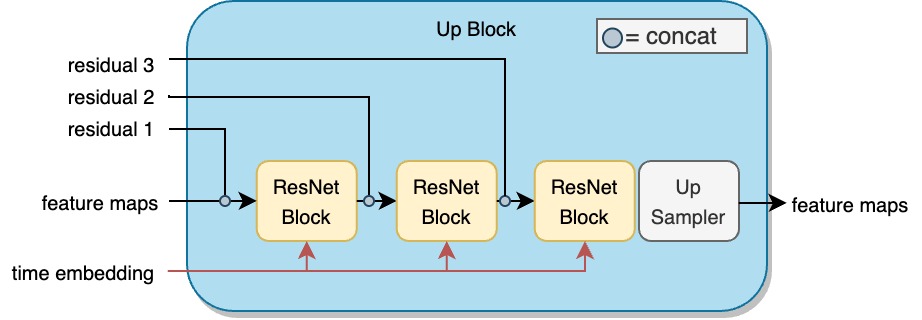}}
    	\end{minipage}
        \hspace{0.01\textwidth}
    	\begin{minipage}{0.49\textwidth}
    	    \centering
            \subfloat[\textbf{Down Block (without attention)} \label{fig:down-block}]{\includegraphics[width=1\textwidth]{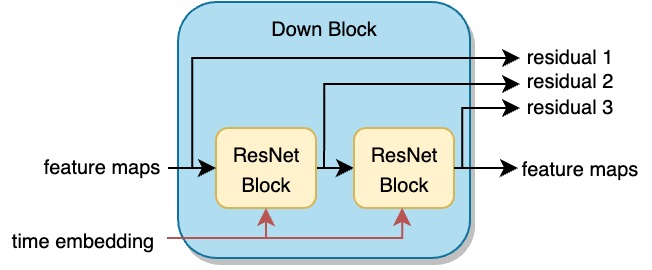}}
    	\end{minipage}
         \begin{minipage}{0.49\textwidth}
    	    \centering
            \subfloat[\textbf{Up Sampler} \label{fig:up-sampler}]{\includegraphics[width=1\textwidth]{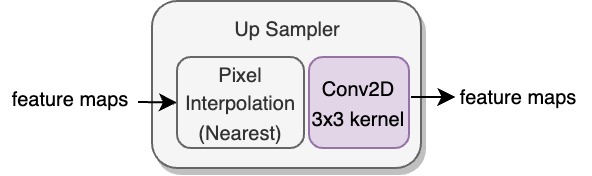}}
    	\end{minipage}%
        \begin{minipage}{0.49\textwidth}
    	    \centering
            \subfloat[\textbf{Cross Attention Mid Block} \label{fig:mid-block}]{\includegraphics[width=1\textwidth]{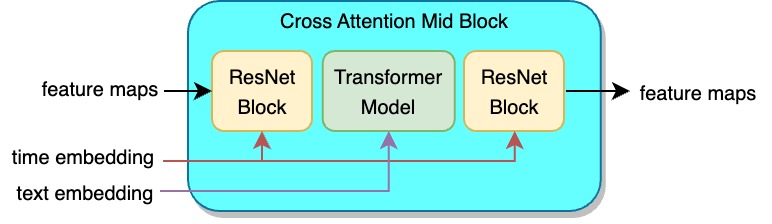}}
    	\end{minipage}%
    
        \caption{Smaller modules from the time-conditional UNet architecture}
        \label{fig:submodules}
    \end{figure}

\begin{figure}[H]
        \centering
        \includegraphics[width=1\textwidth]{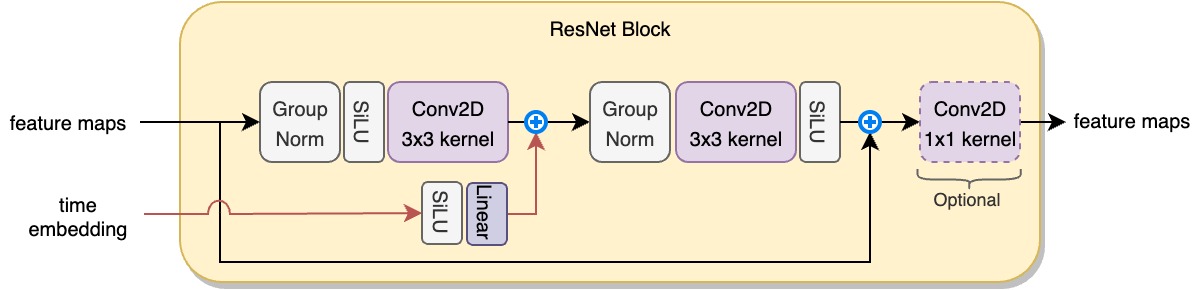}
        \caption{ResNet block submodule from the time-conditional UNet architecture}
        \label{fig:resnet-block}
    \end{figure}

    This network is conditioned on a timestep $t$ using the sinusoidal positional embedding \cite{transformers}. This timestep embedding is projected into the ResNet blocks (Figure \ref{fig:resnet-block}) present in the down and up sampling blocks (see Figures \ref{fig:cross-attention-down-block}, \ref{fig:cross-attention-up-block}, \ref{fig:down-block} and \ref{fig:up-block}). The text conditioning is achieved by tokinization and using CLIP to acquire an embedding which is also injected into the down and up sampling blocks (see Figures \ref{fig:cross-attention-down-block} and \ref{fig:cross-attention-up-block}).

    The ResNet block depicted in Figure \ref{fig:resnet-block} constitutes a sequence of normalisation, activation, and convolutional operations with a residual connection. The time embedding is subjected to a linear layer which serves to transform it to the required dimension before performing a basic element-wise addition. Additionally, the block may include an optional 1x1 kernel convolution to increase or reduce the number of feature maps outputted by the block.

\begin{figure}[H]
        \centering
        \includegraphics[width=1\textwidth]{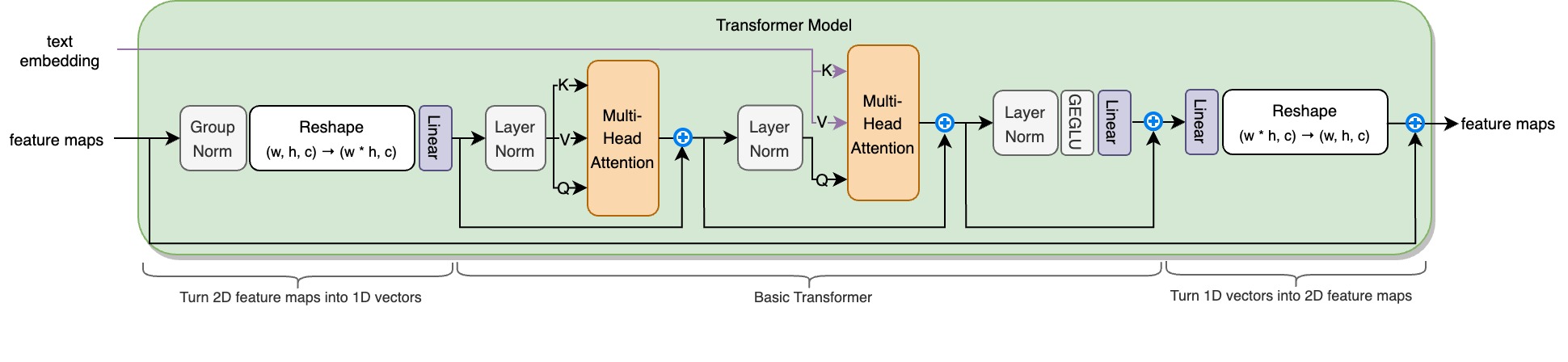}
        \caption{Transformer Model submodule from the time-conditional UNet architecture}
        \label{fig:cross-attention}
    \end{figure}

    The Transformer Model depicted in Figure \ref{fig:cross-attention} can be clearly divided into three distinct stages. Initially, the feature maps undergo a transformation from $\mathbb{R}^{w \times h \times c} \to \mathbb{R}^{w \cdot h \times c}$ to convert the feature maps into one-dimensional vectors. Subsequently, a linear layer is added to alter the number of vectors from $c$ to $n_h \cdot d$, where $n_h$ denotes the number of attention heads and $d$ signifies the dimension of each head. In the second stage, a basic transformer is applied with a self-attention operation on the flattened feature maps and a cross-attention with the text embedding. In the final stage, the 1D vectors are converted back into $c$ two-dimensional feature maps. For more information on the definition of $K$, $V$, and $Q$ for this cross-attention see Section 3.3 in \cite{stable-diffusion}.

    \section{Implementation details}
    \label{appendix:b}

    \subsection{Software \& Hardware Specifications}

    The present paper showcases a series of experiments, all of which were executed on hardware and software configurations that are identical. Additionally, the YOLO models utilised in the experiments were trained using uniform hyperparameters. See Table \ref{table:specs} for an overview.

    \def\arraystretch{1.5}
    \begin{table}[H]
        \begin{minipage}{.5\linewidth}
          \centering
        	\begin{tabular}{cc}
                \multicolumn{2}{c}{\textit{\textbf{\large}}}  \\ 
        		\multicolumn{2}{c}{\textit{\textbf{\large Hardware Specifications}}}  \\  
        		\multicolumn{1}{>{\centering\arraybackslash}p{2.5cm}}{\textbf{Hardware}} &
        		\multicolumn{1}{>{\centering\arraybackslash}p{4cm}}{\textbf{Virtual Machine}} \\ \hline
        		
        		\multicolumn{1}{|>{\centering\arraybackslash}p{2.5cm}|}{CPU} &
        		\multicolumn{1}{>{\centering\arraybackslash}p{4cm}|}{24 Cores @ 2.00GHz}\\ \hline
        		
        		\multicolumn{1}{|>{\centering\arraybackslash}p{2.5cm}|}{RAM} &
        		\multicolumn{1}{>{\centering\arraybackslash}p{4cm}|}{60 GB}\\ \hline
        		
        		\multicolumn{1}{|>{\centering\arraybackslash}p{2.5cm}|}{GPU} &
        		\multicolumn{1}{>{\centering\arraybackslash}p{4cm}|}{NVIDIA A40 \small{(Accessed via a VMware SVGA II Adapter)}}\\ \hline
        		
        		\multicolumn{1}{|>{\centering\arraybackslash}p{2.5cm}|}{GPU Memory} &
        		\multicolumn{1}{>{\centering\arraybackslash}p{4cm}|}{16 GB}\\ \hline
        		
        		\multicolumn{1}{|>{\centering\arraybackslash}p{2.5cm}|}{CUDA version} &
        		\multicolumn{1}{>{\centering\arraybackslash}p{4cm}|}{11.6}\\ \hline
        	\end{tabular}
         
          \centering
        	\begin{tabular}{cc}    
                \multicolumn{2}{c}{\textit{\textbf{\large}}}  \\ 
          		\multicolumn{2}{c}{\textit{\textbf{\large Software Specifications}}}  \\  
        		\multicolumn{1}{>{\centering\arraybackslash}p{2.5cm}}{\textbf{Software}} &
        		\multicolumn{1}{>{\centering\arraybackslash}p{4cm}}{\textbf{Version}} \\ \hline
        		
        		\multicolumn{1}{|>{\centering\arraybackslash}p{2.5cm}|}{PyTorch} &
        		\multicolumn{1}{>{\centering\arraybackslash}p{4cm}|}{1.12.1+cu116}\\ \hline
        		
        		\multicolumn{1}{|>{\centering\arraybackslash}p{2.5cm}|}{Torchvision} &
        		\multicolumn{1}{>{\centering\arraybackslash}p{4cm}|}{0.13.1+cu116}\\ \hline
        		
        		\multicolumn{1}{|>{\centering\arraybackslash}p{2.5cm}|}{Python} &
        		\multicolumn{1}{>{\centering\arraybackslash}p{4cm}|}{3.8.13}\\ \hline
        
        	\end{tabular}
        \end{minipage} 
        \begin{minipage}{.5\linewidth}
          \centering
           	\begin{tabular}{cc}    
                \multicolumn{2}{c}{\textit{\textbf{\large}}}  \\ 
          		\multicolumn{2}{c}{\textit{\textbf{\large YOLO Training Specifications}}}  \\
                \multicolumn{1}{>{\centering\arraybackslash}p{2.5cm}}{\textbf{Parameter}} &
        		\multicolumn{1}{>{\centering\arraybackslash}p{4cm}}{\textbf{Value}} \\ \hline
          
        		\multicolumn{1}{|>{\centering\arraybackslash}p{2.5cm}|}{Batch size} &
        		\multicolumn{1}{|>{\centering\arraybackslash}p{4cm}|}{8} \\ \hline

                \multicolumn{1}{|>{\centering\arraybackslash}p{2.5cm}|}{Learning rate} &
        		\multicolumn{1}{|>{\centering\arraybackslash}p{4cm}|}{$10^{-2}$} \\ \hline

        		\multicolumn{1}{|>{\centering\arraybackslash}p{2.5cm}|}{Testing IoU} &
        		\multicolumn{1}{>{\centering\arraybackslash}p{4cm}|}{0.45}\\ \hline
          
        		\multicolumn{1}{|>{\centering\arraybackslash}p{2.5cm}|}{NMS IoU} &
        		\multicolumn{1}{>{\centering\arraybackslash}p{4cm}|}{0.45}\\ \hline
        		
        		\multicolumn{1}{|>{\centering\arraybackslash}p{2.5cm}|}{Precision} &
        		\multicolumn{1}{>{\centering\arraybackslash}p{4cm}|}{float16 (AMP \cite{mixed-precision})}\\ \hline
        		
        		\multicolumn{1}{|>{\centering\arraybackslash}p{2.5cm}|}{Optimiser} &
        		\multicolumn{1}{>{\centering\arraybackslash}p{4cm}|}{SGD}\\ \hline

                \multicolumn{1}{|>{\centering\arraybackslash}p{2.5cm}|}{Momentum} &
        		\multicolumn{1}{>{\centering\arraybackslash}p{4cm}|}{0.937}\\ \hline
          
                \multicolumn{1}{|>{\centering\arraybackslash}p{2.5cm}|}{Weight decay} &
        		\multicolumn{1}{>{\centering\arraybackslash}p{4cm}|}{$5 \cdot 10^{-4}$}\\\hline
        
        	\end{tabular}
        \end{minipage}%
        \caption{Overview of the hardware and software specifications used during the experiments.}
    	\label{table:specs}
    \end{table}
        
    \subsection{DreamBooth Specifications}

    To effectively run DreamBooth fine-tuning with our limited VRAM, we implemented several optimisations following the guidelines suggested by \cite{github-diffusers}. These optimisations include utilising a memory efficient optimiser \cite{8bitadam} and attention mechanism \cite{xformers}, automatic mixed precision \cite{mixed-precision}, and setting the 'grad' attributes in PyTorch tensors to None instead of 0 when calling $zero\_grad()$.

    \subsubsection{8-bit Optimiser}

    In the field of deep learning, commonly used optimisation techniques include stochastic gradient descent (SGD) with momentum and Adam, both of which involve the tracking of gradient information over time. While the use of past gradient values can accelerate the optimisation process, it comes at the cost of increased memory requirements. Recently, \cite{8bitadam} introduced a new optimisation approach that employs 8-bit statistics for tracking gradient information. This method achieves comparable performance levels to traditional stateful optimisers that rely on 32-bit optimiser states, but with lower memory requirements. To fine-tune the Stable Diffusion models with our limited VRAM, an 8-bit version of the Adam algorithm was employed for this study.

    \subsubsection{Memory-efficient Attention}

    It has been observed that the time and memory complexity of attention usually follows the order of $O(n^2)$. However, in the research article published by \cite{xformers}, they have introduced a straightforward and efficient algorithm that greatly decreases the memory usage for attention. The proposed algorithm provides a method for performing self-attention that requires only $O(\log n)$ memory. Moreover, the authors have demonstrated that the function can be differentiated while preserving its memory efficiency. In this study, we have employed this attention mechanism in the transformer modules from the UNet (see Figure \ref{fig:cross-attention}).
    
    \subsubsection{Hyperparameters}
    
    In reference to hyperparameters, we have set a train batch size of 1 due to our limited VRAM and a learning rate of $2 \cdot 10^{-6}$, with a constant learning rate scheduler. We utilised the Adam optimiser, with $\beta_1 = 0.9$ and $\beta_1 = 0.999$, with a weight decay of $10^{-2}$ and an epsilon value of $10^{-8}$. Furthermore, we configured the $\lambda$ value for the prior preservation regularisation to be $1.0$. 
    
    In terms of training, we have used 6 epochs. However, for the annotation generation fine-tuning, as described in Appendix \ref{appendix:e}, we have determined that a minimum of 12 epochs is necessary. We hypothesize that the network requires more time to "unlearn" the information that is typically present in the blue channel and learn to associate it with our encoded annotations.

    \section{Generated Images}
    \label{appendix:c}

    \begin{figure}[H]
        \centering
        \includegraphics[width=0.93\textwidth]{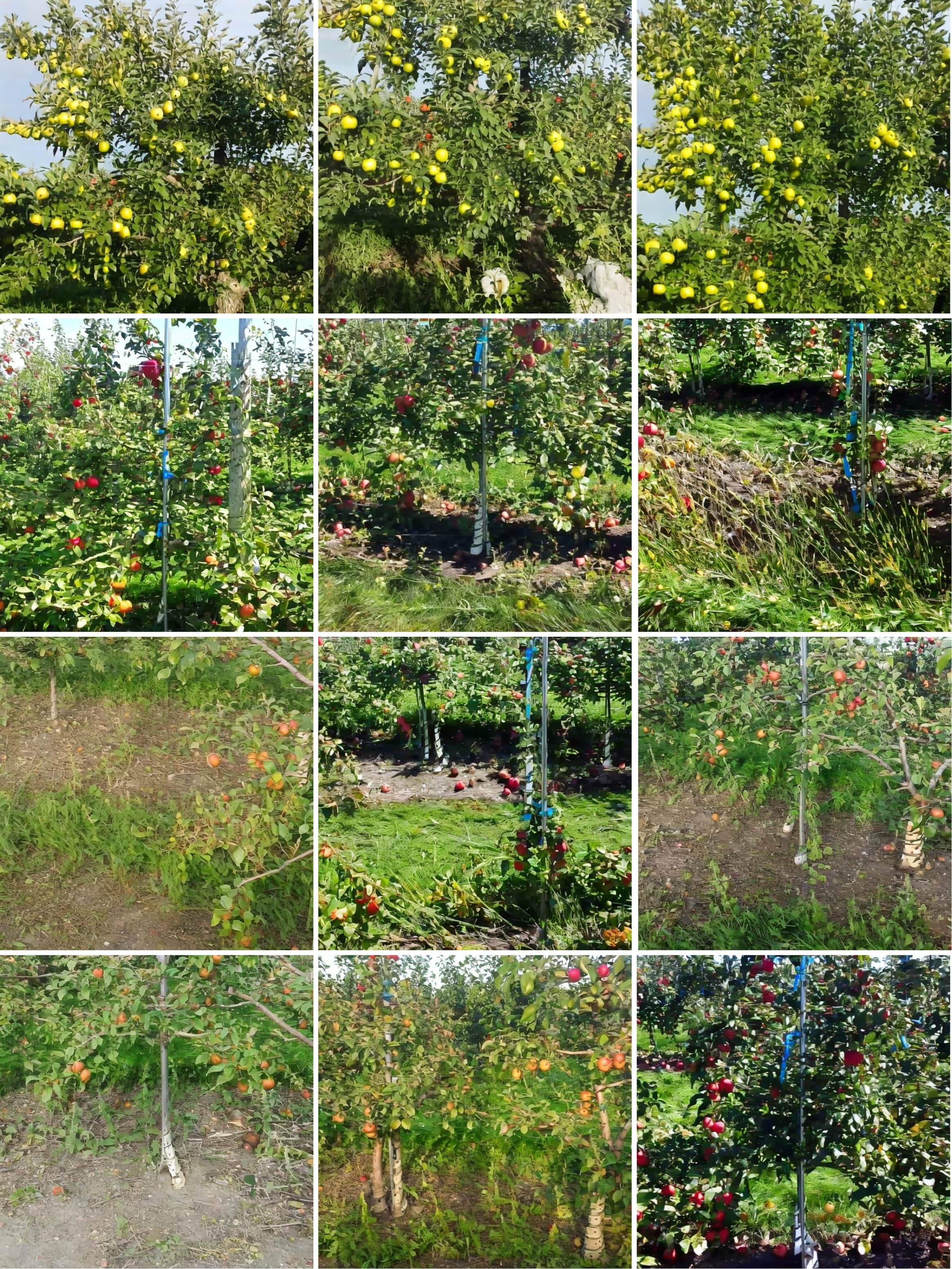}
        \caption{A bigger subset of our generated images from our Genfusion Pipeline.}
        \label{fig:generated-tree-images}
    \end{figure}

    \section{Instance Images \& Class Images}
    \label{appendix:d}

    \begin{figure}[H]
        \centering
        \includegraphics[width=0.9\textwidth]{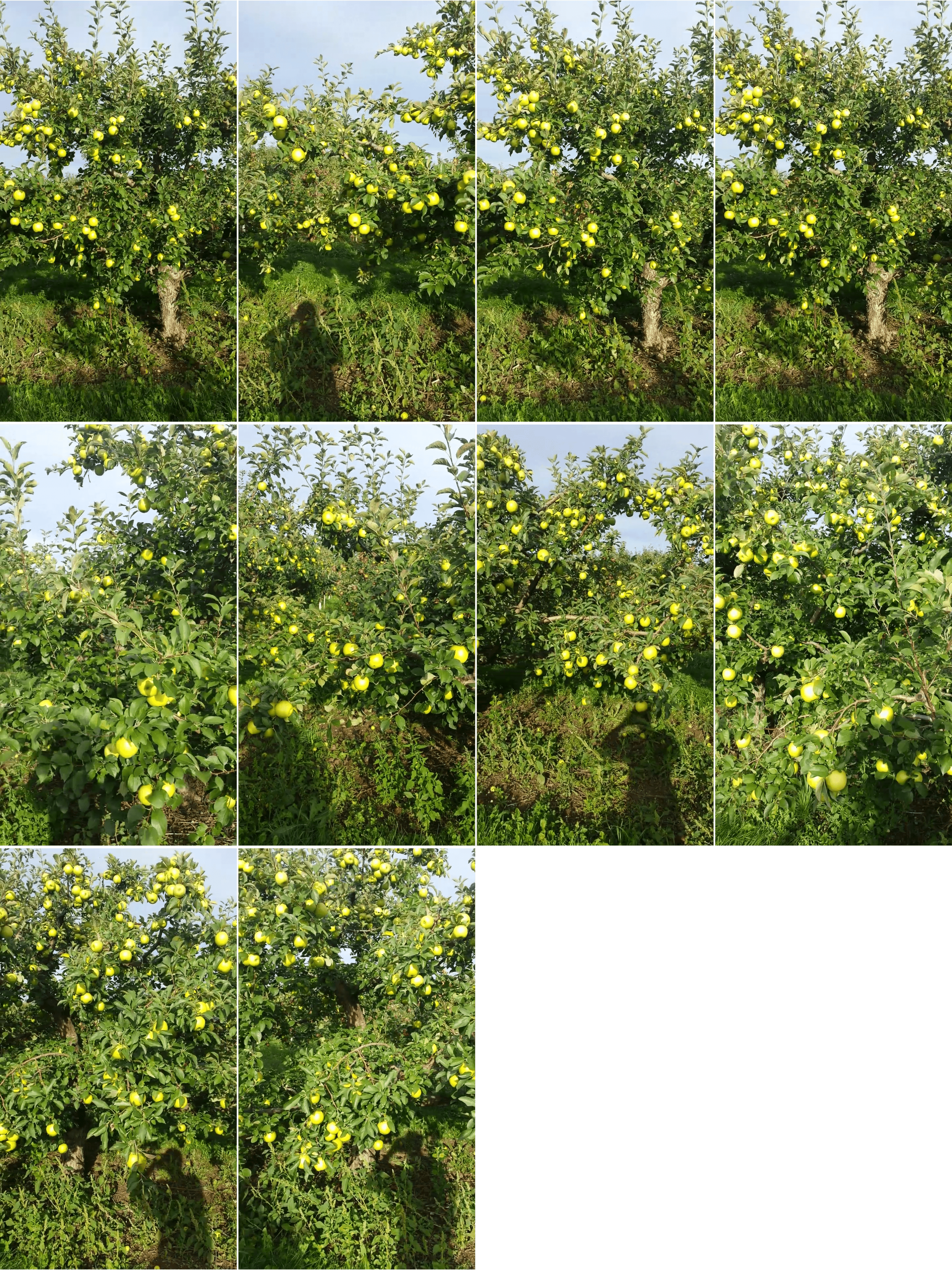}
        \caption{The 10 green apple tree images sampled from the MinneApple \cite{hani2020minneapple} train set. $\mathcal{X}$ is created from these samples by creating crops of 720x720 and then pixel interpolating them to match the output size of Stable Diffusion (768x768).}
        \label{fig:instance-images-green}
    \end{figure}

    \begin{figure}[H]
        \centering
        \includegraphics[width=1\textwidth]{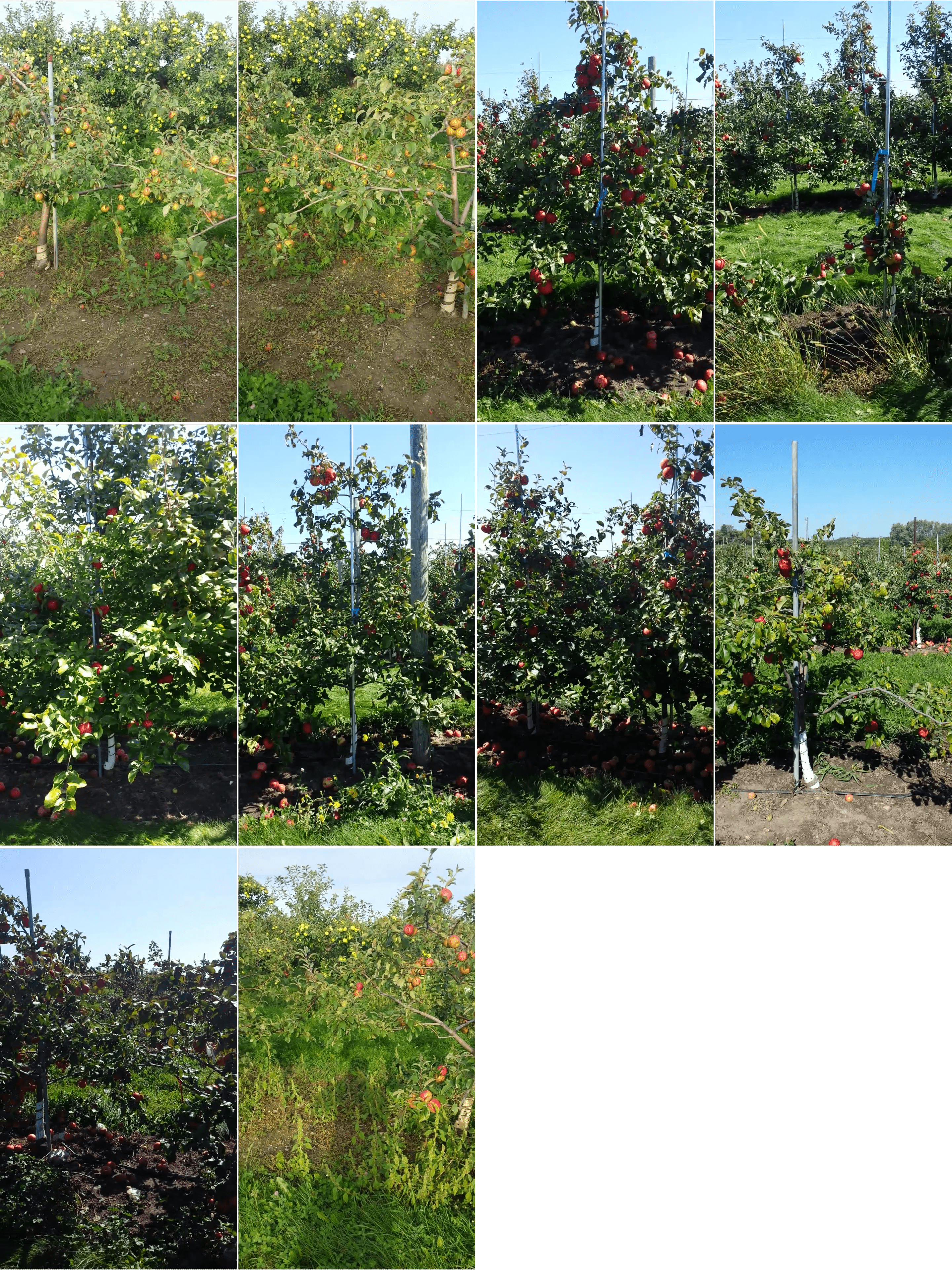}
        \caption{The 10 red apple tree images sampled from the MinneApple \cite{hani2020minneapple} train set. $\mathcal{X}$ is created from these samples by creating crops of 720x720 and then pixel interpolating them to match the output size of Stable Diffusion (768x768).}
        \label{fig:instance-images-red}
    \end{figure}

    \begin{figure}[H]
        \centering
        \includegraphics[width=1\textwidth]{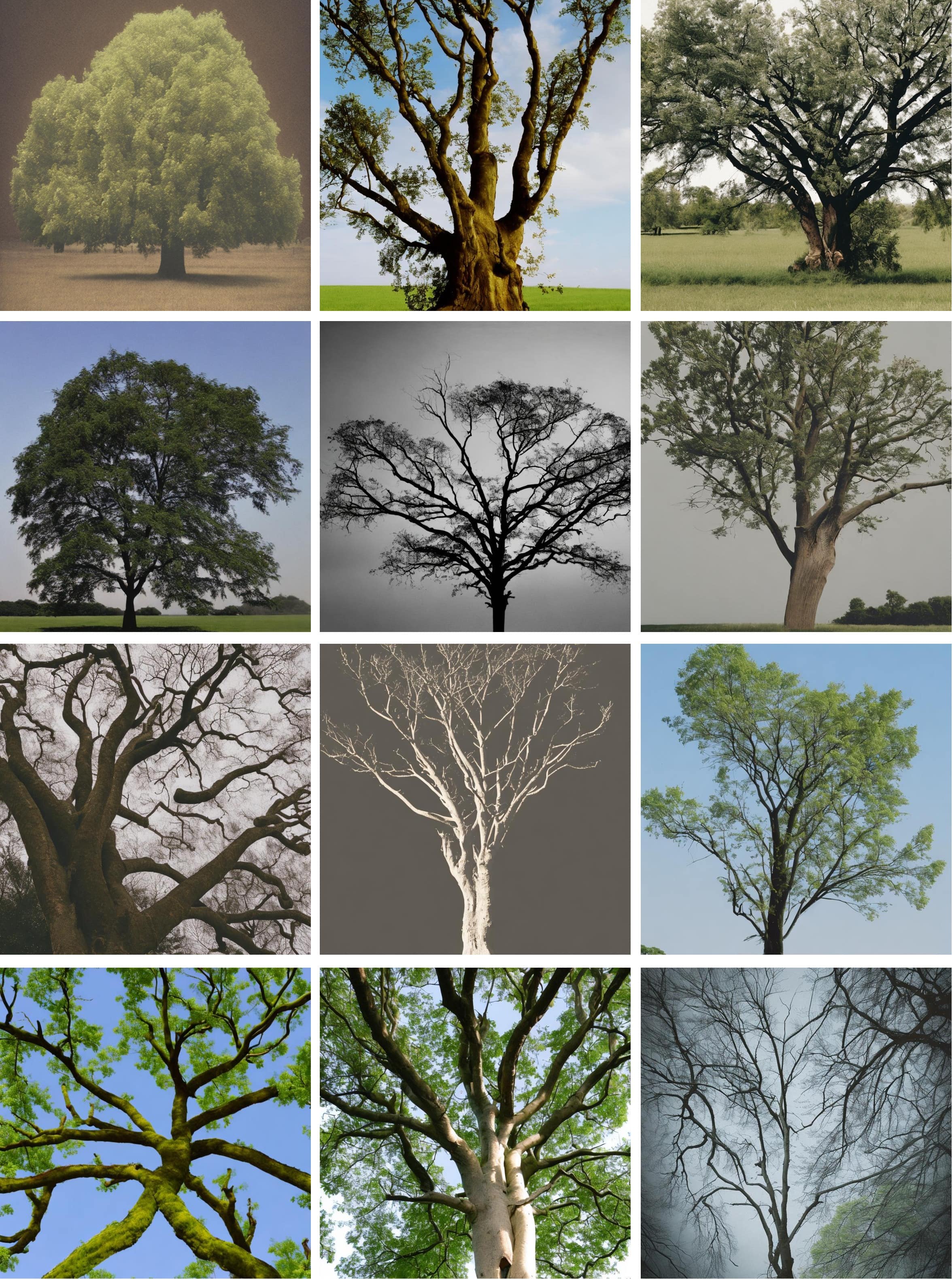}
        \caption{A subset of the $\mathcal{X}^{(pr)}$ class specific images generated with the prompt "A tree" before fine-tuning.}
        \label{fig:class-images}
    \end{figure}

    \begin{figure}[H]
        \centering
        \includegraphics[width=1\textwidth]{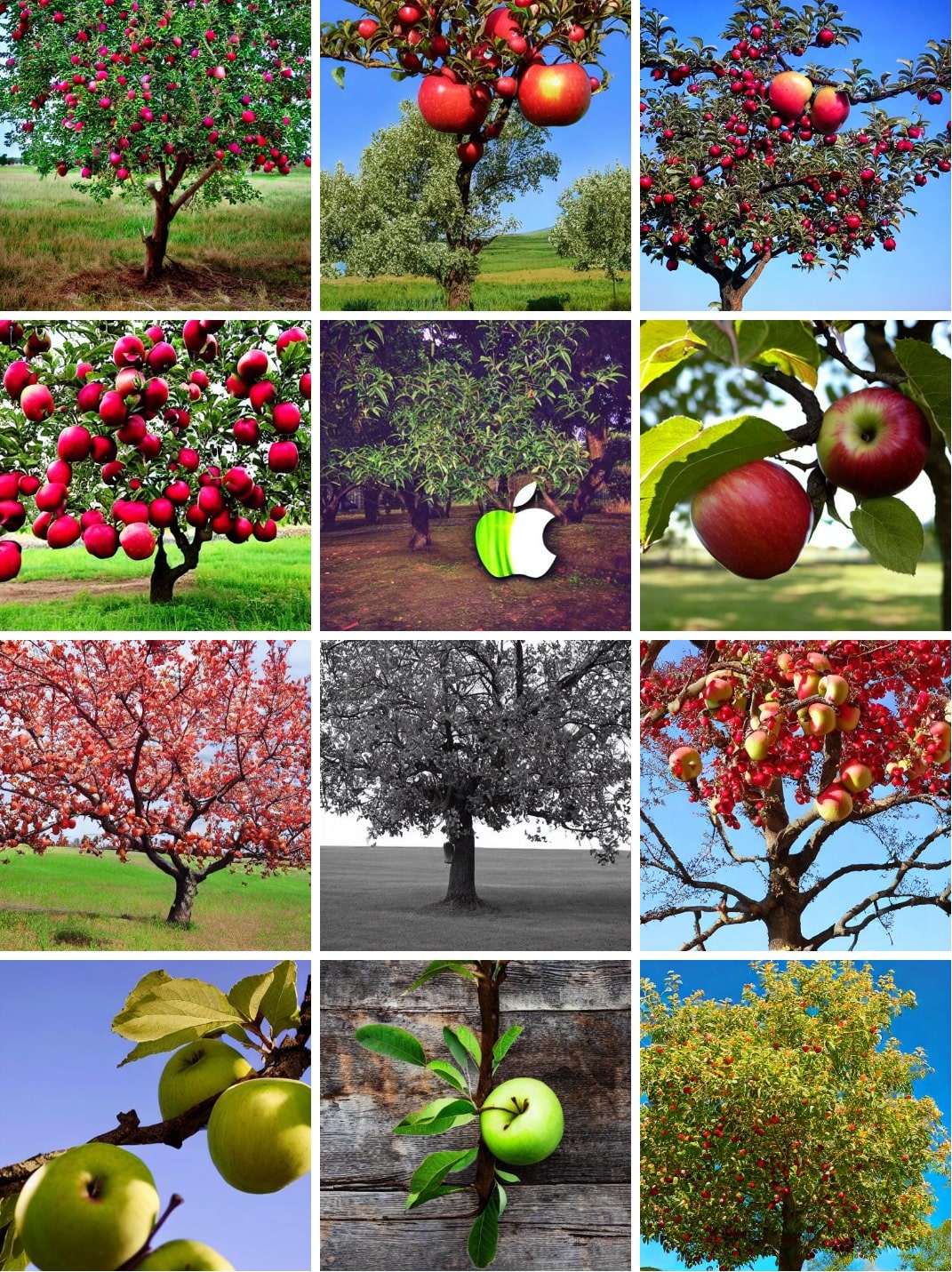}
        \caption{The rationale behind selecting the prompt "a tree" instead of "an apple tree" was based on the non-realistic images generated with the "an apple tree" prompt.}
        \label{fig:apple-tree-images}
    \end{figure}

    \section{Preliminary Investigation into Automatic Annotation Generation}
    \label{appendix:e}

    Given the impressive performance of diffusion models in generating image data, we explored the potential of encoding annotation data within the image data. Initially, our approach involved introducing a fourth channel to encode object locations. However, we encountered a challenge when utilising the VQ-VAE from latent diffusion, as it could only process 3D inputs. To overcome this limitation, we repurposed the blue channel as the annotation channel. This decision was based on the hypothesis that the blue information would be less critical for detecting green and red apples. Additionally, considering the scope of our study, undertaking a retraining process for the VQ-VAE was not feasible.

\begin{figure}[H]
        \centering
        \includegraphics[width=1\textwidth]{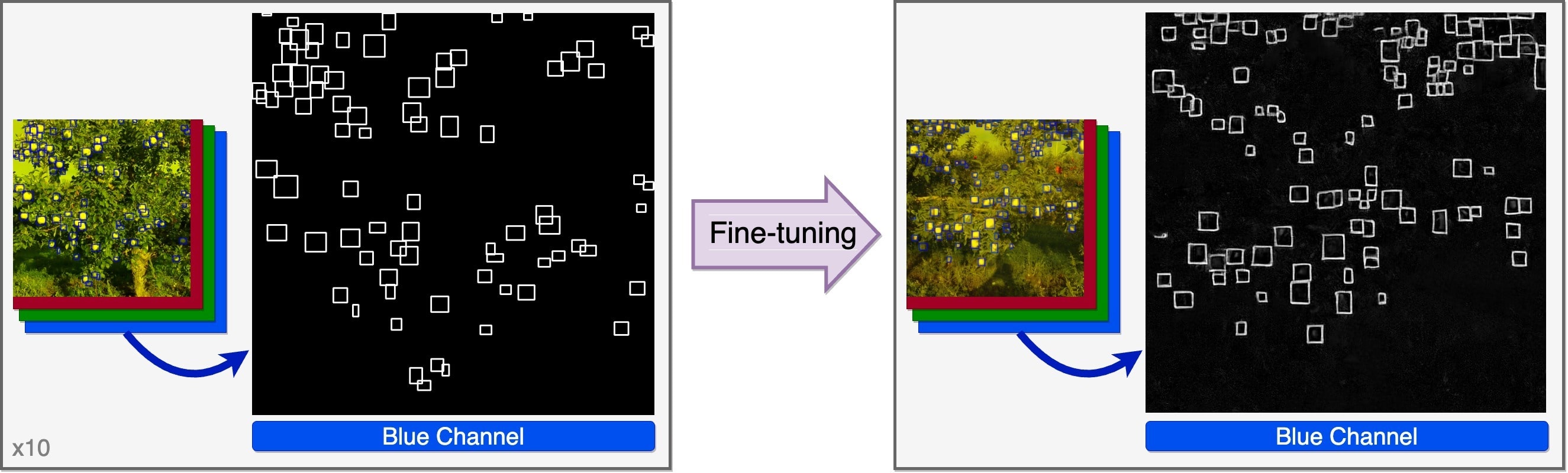}
        \caption{\textbf{Fine-tuning with bounding box outlines.} The input images (left) corresponded to the same MinneApple images as those used for the generator presented in the main paper. However, in this case, the blue channel information was substituted with bounding box outlines. Upon fine-tuning, the resulting images (right) showed promising outcomes, as the blue channel featured proper annotations upon initial inspection.}
        \label{fig:bbox-annot}
    \end{figure}

    Our initial experiment involved the encoding of annotation information as the outlines of bounding boxes, as depicted on the left side of Figure \ref{fig:bbox-annot}. The right side of the figure demonstrates the results obtained from the fine-tuning process conducted on images with two colour channels, along with the bounding box outline in the blue channel. Notably, the fine-tuning procedure for these images required approximately double the number of epochs compared to fine-tuning on regular 3-colour channel images in order to achieve successful results. We hypothesise that the pretrained diffusion model initially needs to undergo a process of 'unlearning' the conventional understanding of the content typically associated with the blue channel.
    
    Although these results appear promising regarding accurate marking of the apples in the images, when wanting to extract bounding box coordinates challenges arise when confronted with scenarios involving overlapping bounding boxes. To overcome this limitation, we considered encoding the annotations as instance segmentation masks. However, this method also presented its own limitations. Specifically, it would necessitate retraining the VQ-VAE from latent diffusion to preserve subtle pixel intensity differences in the blue channel after encoding and decoding. Unfortunately, given the scope of our study, undertaking such a retraining process was deemed infeasible.

\begin{figure}[H]
        \centering
        \includegraphics[width=1\textwidth]{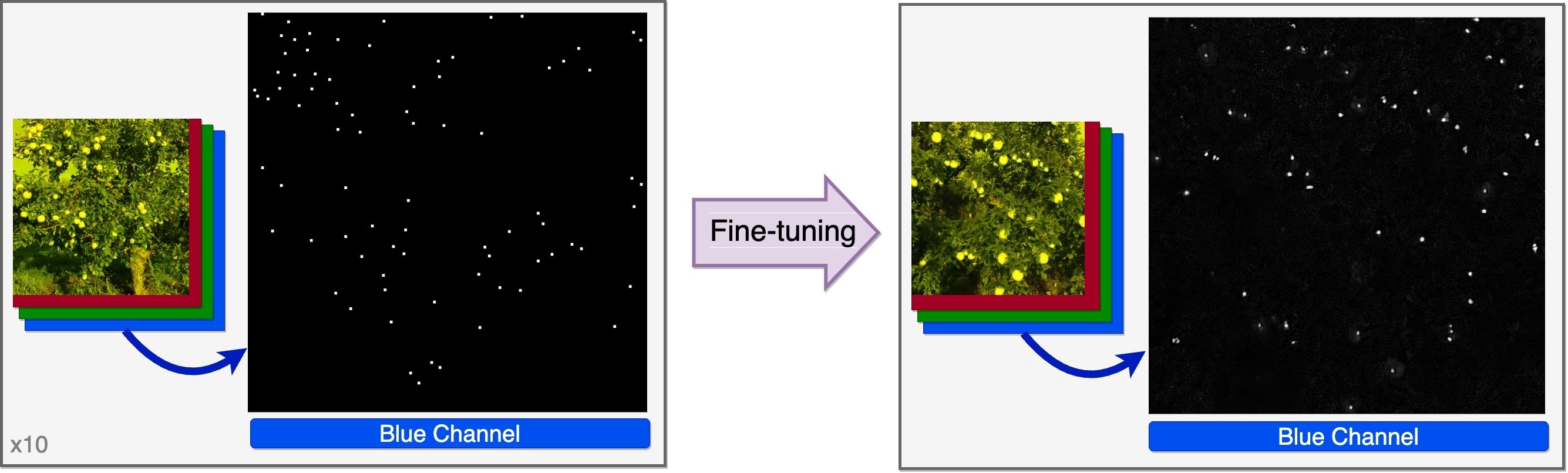}
        \caption{\textbf{Fine-tuning with dot annotations.} This is the same fine-tuning as in the previous figure on bounding box outlines with input images (left) being from MinneApples and output images (right) showing promising results when inspecting the annotation (blue) channel.}
        \label{fig:dot-annot}
    \end{figure}

    Subsequently, we explored an encoding approach utilising high-intensity dots positioned at the centre of each object, as illustrated on the left in Figure \ref{fig:dot-annot}. With this dot-based encoding method, annotations could be easily extracted using "find contours" and "image moments," offering notable advantages in scenarios with overlapping objects. The right side of Figure \ref{fig:dot-annot} illustrates that, once again, the diffusion model was capable of generating accurate information in the blue channel. However, it is important to acknowledge that this encoding technique does not preserve the width and height information of the objects. To address this limitation, recent studies, such as Segment Anything \cite{sam}, have demonstrated promising results by utilising dots as guides to create masks that accurately represent the object's width and height at the corresponding image location. Notably, in the dot-based annotation encoding approach, we observed that the high-intensity areas needed to cover a minimum of 5x5 pixels to achieve satisfactory outcomes. Additionally, we observed that if the pixel areas were smaller, these pixel dots were automatically removed by the VQ-VAE.

    Upon further investigation, we discovered an increased occurrence of generated images did not bear any resemblance to the images from MinneApple and instead depicted regular trees. To address this issue, we employed several postprocessing techniques. Firstly, we performed Principal Component Analysis (PCA) on all the images to reduce them to 50 dimensions. Subsequently, we applied t-Distributed Stochastic Neighbor Embedding (t-SNE) dimensionality reduction to further reduce them to 2 dimensions. Finally, we applied K-Means clustering with a cluster count of 3 and retained only the images in the largest cluster. Consequently, we created a dataset similar to the one described in the main paper, containing 54 green apple trees and 842 red apple trees.

\begin{figure}[H]
        \centering
        \includegraphics[width=0.6\textwidth]{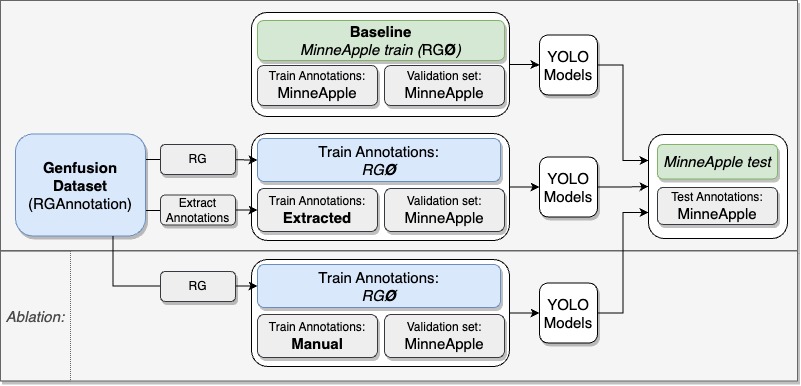}
        \caption{Experiment Setup}
        \label{fig:experiment-2}
    \end{figure}

    With this newly created dataset, we devised the following experimental plan to evaluate the accuracy of our approach, as depicted in Figure \ref{fig:experiment-2}. In this figure, we refer to an image with three colour channels as "RGB", an image with an empty blue channel as "RG$\varnothing$", and an image with encoded annotations in the blue channel as "RGAnnotation". As a baseline, we wanted to train an object detector solely on the red and blue colour information to test our hypothesis regarding the lesser significance of the blue channel for our detection task. Subsequently, with our generated dataset we wanted to train a detector on both the extracted annotations and, as an ablation, on manually annotated data. This approach would allow us to ascertain the performance drop attributable to the absence of the blue channel, the utilisation of generated data, and the use of our extracted annotations.

    Now with this dataset created, we designed the following experiment plan to test the accuracy of our approach, see Figure \ref{fig:experiment-2}. In this figure, we refer to an image with three colour channels as "RGB", an image with an empty blue channel as "RG$\varnothing$", and an image with encoded annotations in the third channel as "RGAnnotation". As a baseline, we trained an object detector on only the red and blue colour information to test our hypothesis of the blue channel being of less significance for our detection task. Then we wanted to create a dataset using our dot-based annotation generation technique and train a detector both on our extracted annotations and as an ablation on manual annotations. In this way we would be certain what drop in performance was due to the dropping of the blue channel, what drop in performance was due to the use of generated data, and what drop in performance was due to our extracted annotations. Unfortunately, we never finished this plan in its entirety because we found one vital flaw when training on the generated data.

\begin{figure}[H]
        \centering
        \includegraphics[width=0.30\textwidth]{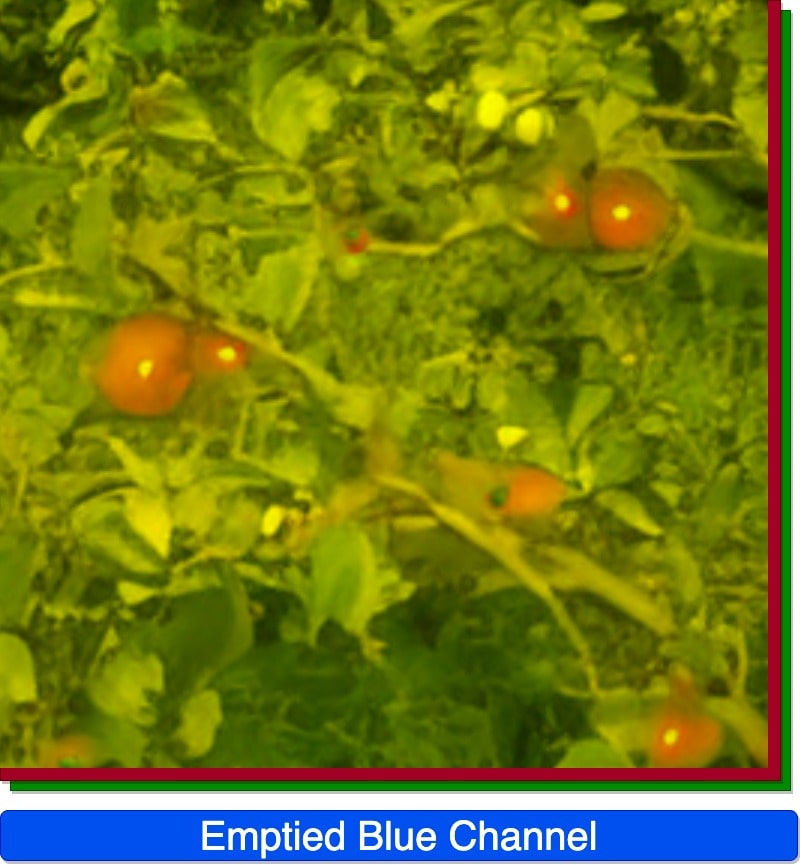}
        \caption{Zoomed in depiction of the red and green colour channels of a generated image revealing the discussed 'channel leakage'.}
        \label{fig:wrong-annot}
    \end{figure}

    While conducting our analysis, we discovered that a model trained on MinneApple RGB achieved an AP@0.5:0.05:0.95 score of 0.43, whereas a model trained on RG$\varnothing$ obtained a score of 0.37, resulting in a marginal AP difference of only 0.06. However, as we transitioned towards utilising the generated data, we encountered a significant limitation that affected our findings. Specifically, we observed that the information from the annotation channel exhibited a phenomenon of 'leakage' to the other two channels, as illustrated in Figure \ref{fig:wrong-annot}. This leakage manifested as the presence of high-intensity dots at the centre of apples in the red and green channels, leading to severe overfitting even when training the model exclusively on the green and red colour channels. In an attempt to mitigate this overfitting issue, we explored various hyperparameter adjustments, including lower learning rates, switching from SGD to Adam, and experimenting with models of different sizes ranging from medium to nano. Despite these efforts, we were unable to achieve an AP metric surpassing the 0.10 score.

    Unfortunately, we did not have the resources to further study this avenue of research, however, our results do show the versatility this DreamBooth fine-tuning has. Based on the findings and limitations identified in our experiments, there are several options for future research. The following areas deserve further exploration and investigation:

    \begin{enumerate}
        \item \textbf{Leakage Mitigation:} The issue of information leakage from the annotation channel to other colour channels poses a significant challenge in training models solely on the green and red colour channels. Future work should focus on developing techniques to effectively address this leakage problem, such as advanced regularisation methods or novel network architectures specifically designed to handle multi-channel annotations.

        \item \textbf{Enhanced Encoding Techniques:} While the dot-based encoding approach demonstrated advantages in scenarios with overlapping objects, it fell short in preserving the width and height information of the objects. Building upon recent studies like Segment Anything \cite{sam}, future research could investigate the development of innovative encoding techniques that accurately capture both the spatial location and the size attributes of objects in a more comprehensive manner.
    \end{enumerate}

\end{appendices}

\end{document}